\documentclass{article}
\usepackage{arxiv}

\usepackage{geometry}
\usepackage{xcolor}
\usepackage{hyperref}
\usepackage{titlesec}
\usepackage{enumitem}
\usepackage{amsmath}
\usepackage{amssymb}
\usepackage{graphicx}
\usepackage{booktabs}
\usepackage{cite}
\usepackage{array}
\newcolumntype{C}[1]{>{\centering\arraybackslash}p{#1}}
\usepackage{xcolor}
\usepackage{colortbl}
\usepackage{comment}
\usepackage{subcaption}
\usepackage{makecell}
\usepackage{arydshln}
\usepackage{pifont}
\usepackage{xcolor}
\newcommand{\cmark}{\ding{51}}
\newcommand{\xmark}{\ding{55}}
\usepackage{booktabs}      
\usepackage{multirow}      
\usepackage{siunitx}       
\usepackage{makecell}      
\usepackage{amsmath}       
\usepackage{multirow}
\usepackage{siunitx}      
\definecolor{alrow}{RGB}{220, 230, 242}
\usepackage{pifont}
\definecolor{headerblue}{RGB}{220, 230, 242}
\definecolor{algreen}{RGB}{214, 234, 214}
\definecolor{improvegreen}{RGB}{0, 140, 0}
\definecolor{worsenred}{RGB}{180, 0, 0}
\definecolor{neutralgray}{RGB}{100, 100, 100}
\definecolor{bestrow}{RGB}{255, 243, 205}   

\definecolor{feedsblue}{RGB}{30, 80, 160}

\title{%
  \textbf{LoRe-MIM: Lost-or-Replaced Masked Image Modeling for Self-Supervised Representation Learning in Whole-Body PET/CT}\\
}

\author{ 
    Amartya Bhattacharya\textsuperscript{*}  \\
Department of Biomedical Data Science\\
Geisel School of Medicine at Dartmouth\\
Hanover, NH 03755, USA\\
	\texttt{Amartya.Bhattacharya.GR@dartmouth.edu} \\
	\And
    Bashirul Azam Biswas\textsuperscript{*}  \\
Department of Biomedical Data Science\\
Geisel School of Medicine at Dartmouth\\
Hanover, NH 03755, USA\\
	\texttt{Bashirul.Azam.Biswas@dartmouth.edu} \\
    \And
	Biratal Raj Wagle \\
Department of Biomedical Data Science\\
Geisel School of Medicine at Dartmouth\\
Hanover, NH 03755, USA\\
	\texttt{Biratal.Raj.Wagle@dartmouth.edu} \\
    \And
    Michael S. Leapman \\
Department of Urology\\
Yale University\\
New Haven, CT, USA\\
\texttt{michael.leapman@yale.edu}\\ 
    \And
	Matthew E. Maeder \\
Radiology\\
Dartmouth Hitchcock Medical Center\\
Lebanon, NH 03766 , USA\\
\texttt{Matthew.E.Maeder@dartmouth.edu}\\ 
    \And
    James B. Yu \\
Radiation Oncology\\
Dartmouth Hitchcock Medical Center\\
Lebanon, NH 03766 , USA\\
	\texttt{James.B.Yu@dartmouth.edu}\\ 
    \And
	Indrani Bhattacharya \\
Department of Biomedical Data Science\\
Geisel School of Medicine at Dartmouth\\
Hanover, NH 03755, USA\\
	\texttt{Indrani.Bhattacharya@dartmouth.edu}\\
}
\date{}

\renewcommand{\shorttitle}{LoRe-MIM}

\begin{document}

\maketitle

\begingroup
\renewcommand{\thefootnote}{*}
\footnotetext{These authors contributed equally to this work.}
\endgroup

\begin{abstract}
Deep learning models for whole-body positron emission tomography/computed tomography (PET/CT) hold significant clinical potential for disease detection, treatment planning, and response assessment. However, the substantial time and cost associated with expert annotations remain major bottlenecks in developing robust and generalizable models that can assist with PET/CT interpretation. Although self-supervised learning (SSL) has emerged as a promising approach to mitigating annotation scarcity in natural and medical imaging, its application to whole-body PET/CT remains relatively underexplored due to the limited availability of large-scale datasets and the challenges associated with learning complementary cross-modal representations. To address these limitations, we curate a heterogeneous, multi-tracer, multi-institutional cohort comprising 6,331 whole-body PET/CT scans and propose \textbf{LoRe-MIM (Lost-Or-Replaced Masked Image Modelling)}, a novel self-supervised framework designed to learn multimodal representations through masked image reconstruction. Specifically, \textbf{LoRe-MIM} introduces a masking strategy in which image regions are either replaced with zeros (\textbf{Lost}) or substituted with corresponding intensity values from the complementary imaging modality (\textbf{Replaced}), encouraging the model to capture both modality-specific characteristics and cross-modal relationships. We systematically evaluate the learned representations across multiple downstream tasks, including multimodal PET/CT image reconstruction, patch-based classification into cancer vs. non-cancer, whole-body lesion segmentation, label-efficient training, and robustness assessment to noise, missing modality, and PET/CT misalignment. We assess generalizability in performance across tracers and different institutional datasets. Experimental results demonstrate that LoRe-MIM consistently outperforms the corresponding state-of-the-art baselines across these tasks, highlighting its potential as a scalable and generalizable self-supervised learning framework for whole-body PET/CT analysis.
\end{abstract}

\section{Introduction}\label{sec:intro}

Cancerous lesions in whole-body imaging exhibit substantial morphological variability in their physical appearance, shape, size, and structural characteristics across patients, cancer types, and anatomical sites~\cite{hofman2016we,mingels2025total,belal2024applications}. Noninvasive imaging technologies such as positron emission tomography/computed tomography (PET/CT) play a critical role in detecting such lesions, determining disease stage, guiding treatment planning, and assessing treatment response~\cite{trotter2023positron}. PET captures the molecular and physiological activity of lesions, whereas CT provides complementary anatomical and structural context. However, with the rising global cancer burden~\cite{sung2026global}, the volume of whole-body PET/CT examinations is increasing at a substantially faster pace than the availability of expert nuclear medicine specialists capable of interpreting them. This growing imbalance creates a critical need for automated systems that can efficiently and accurately analyze large volumes of whole-body PET/CT scans and assist clinicians. As in many other medical imaging domains, deep-learning-based methods have emerged as a promising approach for automating oncologic whole-body PET/CT interpretation~\cite{gatidis2022fdg,ingrisch2024autopet,rokuss2024fdg}.

Deep-learning-based models have been developed for whole-body lesion segmentation~\cite{xu2018automated, blanc2021fully, jemaa2020tumor, he2023whole, gatidis2024results, liu2024lm, xu2023automatic, li2024automated, milletari2016v} and patch-level  classifications tasks (cancer presence or absence)~\cite{sibille202018f,mu2020non,shao2021application,alves2024classification,tao202118f,yang2023computer,alves2024classification,shao2024transfer,nishigaki2024vision}. The accuracy and generalizability of deep learning methods for both tasks depend heavily on the ability of the learned PET/CT imaging representations to distinguish between cancer and non-cancer voxels, capture diversity and heterogeneity in imaging features across patient populations, cancer types, tracers, scanners, and anatomical sites, and the robustness to noise, registration errors, and missing modality. Two straightforward approaches to learning such representations are supervised training with task-specific annotations and transfer learning from a model pretrained on a related domain.

Supervised deep learning approaches have been extensively investigated for lesion segmentation in whole-body PET/CT imaging~\cite{xu2018automated, blanc2021fully, jemaa2020tumor, he2023whole, gatidis2024results, liu2024lm, xu2023automatic, li2024automated, milletari2016v}. Blanc-Durand \textit{et al.}~\cite{blanc2021fully} employed a 3D U-Net architecture for the automated detection and segmentation of lymphoma lesions in whole-body FDG-PET/CT scans. Jemaa \textit{et al.}~\cite{jemaa2020tumor} introduced a multi-stage segmentation framework combining 2D segmentation, connected-component analysis, and 3D segmentation-based refinement to improve lesion delineation in whole-body FDG-PET/CT images. To facilitate advancements in whole-body tumor segmentation, Gatidis \textit{et al.}~\cite{gatidis2024results} organized the AutoPET-II challenge and made publicly available a collection of whole-body FDG-PET/CT scans representing diverse tumor types. Building upon this dataset, He \textit{et al.}~\cite{he2023whole} incorporated camouflaged object detection strategies, while Liu \textit{et al.}~\cite{liu2024lm} developed hybrid Mamba-CNN architectures for automated lesion detection and segmentation. Although FDG remains the predominant radiotracer in deep learning-based whole-body PET/CT segmentation research, alternative tracers, including $^{68}$Ga-Pentixafor~\cite{xu2018automated} and prostate-specific membrane antigen (PSMA)-targeted tracers~\cite{xu2023automatic,li2024automated}, have also been investigated in some studies. Specifically, Xu \textit{et al.}~\cite{xu2018automated} designed a cascaded W-Net architecture comprising two V-Nets~\cite{milletari2016v}, where the first network processes CT images and the second integrates $^{68}$Ga-Pentixafor PET with CT information for whole-body multiple myeloma detection. For whole-body PSMA-PET/CT lesion segmentation, Li \textit{et al.}~\cite{li2024automated} proposed a hybrid 3D Transformer–CNN framework, whereas Xu \textit{et al.}~\cite{xu2023automatic} employed U-Net-based architectures. Collectively, these supervised PET/CT segmentation approaches leverage whole-body imaging to identify and delineate lesions associated with single or multiple tumor types.

Similar to lesion segmentation, earlier works on lesion classification relied on supervised training of convolutional neural networks (CNNs) \cite{sibille202018f,mu2020non,shao2021application,alves2024classification}, whereas several recent works \cite{tao202118f,yang2023computer,alves2024classification,shao2024transfer,nishigaki2024vision} have explored transfer learning from pretrained models. Sibille \textit{et al.} proposed a CNN to localize foci of increased FDG uptake and classify lesions suspicious for lung cancer~\cite{sibille202018f}. Mu \textit{et al.} developed a supervised convolutional network using FDG PET/CT scans to classify epidermal growth factor receptor (EGFR) mutation status in patients with non-small cell lung cancer (NSCLC)~\cite{mu2020non}. Shao \textit{et al.} trained a dual-stream 3D CNN using $^{18}$F-FDG PET/CT to distinguish benign and invasive adenocarcinoma in ground-glass lung nodules~\cite{shao2021application}. Alves \textit{et al.} investigated three 3D CNN architectures inspired by the classical 2D architectures AlexNet~\cite{krizhevsky2012imagenet}, VGGNet~\cite{simonyan2014very}, and Inception-v2~\cite{szegedy2016rethinking} for pulmonary nodule classification using $^{18}$F-FDG PET/CT images~\cite{alves2024classification}. They also investigated transfer learning from ResNet-50~\cite{he2016deep} pretrained on ImageNet~\cite{deng2009imagenet}. Tao \textit{et al.} proposed an ensemble ResNet-based classification framework that leveraged transfer learning to extract features from CT, PET, and fused PET/CT images, integrating individual classifiers through majority voting for lung tumor diagnosis~\cite{tao202118f}. Yang \textit{et al.} investigated six ImageNet-pretrained CNN architectures for distinguishing metastatic and lymphomatous cervical lymph nodes using PET/CT images, employing both fine-tuned CNN classifiers and pretrained deep feature extractors combined with support vector machines~\cite{yang2023computer}. Nishigaki \textit{et al.} fine-tuned a Vision Transformer (ViT) pretrained on ImageNet-21k to classify benign and malignant slices from $^{18}$F-FDG PET/CT scans, demonstrating superior performance compared with pretrained CNN architectures, including DenseNet and EfficientNet~\cite{nishigaki2024vision}. Recognizing the limitations of transferring representations learned from 2D natural images to 3D medical imaging, Shao \textit{et al.} explored transfer learning from Models Genesis for predicting EGFR mutation status in NSCLC patients~\cite{shao2024transfer}. 

As observed in existing whole-body PET/CT studies, lesion segmentation is commonly formulated as a fully supervised dense prediction task, whereas lesion classification often leverages representations transferred from models pretrained on large external datasets such as ImageNet or natural-image foundation models. Supervised learning, however, is constrained by the difficulty of constructing sufficiently large and diverse annotated datasets, as reliable lesion-level and voxel-level annotations are expensive to obtain and require substantial sub-speciality clinical expertise. Transfer learning from the 2D natural-image domain to 3D medical imaging introduces a different limitation: a substantial domain mismatch arising from differences in image dimensionality, visual appearance, intensity distributions, and underlying anatomical and functional characteristics.

Self-supervised pretraining offers a unified alternative by exploiting large collections of unlabeled PET/CT scans to learn domain-specific representations that can subsequently be adapted to segmentation through fine-tuning or used as feature extractors for lesion classification. In recent years, several studies have explored self-supervised pretraining for whole-body PET/CT lesion analysis~\cite{yazdani2024automated,oh2025fratmae,huang2025c2maot,keshavarz2026self,liu2026open}. Yazdani \textit{et al.} adopted self-supervised SwinUNETR pretraining using 752 whole-body $[^{68}\mathrm{Ga}]$-PSMA-11 PET/CT scans for prostate cancer metastasis segmentation and demonstrated the benefit of pretraining for downstream lesion delineation~\cite{yazdani2024automated}. Oh \textit{et al.} applied contrastive-learning-based pretraining using 1,292 samples from the publicly available AutoPET III dataset and evaluated its benefit for both lesion segmentation and staging classification on a private institutional cohort~\cite{oh2025fratmae}. Huang \textit{et al.} proposed a modality-complementary masking strategy to learn cross-modal PET/CT representations using 507 FDG cases from AutoPET III~\cite{huang2025c2maot}. Liu \textit{et al.} similarly adopted modality-complementary masking for an FDG-focused foundation model pretrained on 4,997 scans~\cite{liu2026open}. Keshavarz \textit{et al.} employed multimodal SwinUNETR pretraining using 717 $[^{18}\mathrm{F}]$FDG PET/CT scans for downstream segmentation of diffuse large B-cell lymphoma (DLBCL)~\cite{keshavarz2026self}. Collectively, these studies demonstrate the potential of self-supervised pretraining to alleviate annotation scarcity in whole-body PET/CT imaging. However, current self-supervised pretraining approaches remain limited in the scale and diversity of pretraining data, cross-modal interaction modeling, breadth and clinical comprehensiveness of downstream evaluation as detailed below.

\begin{enumerate}
    \item \textbf{Limited pretraining scale and tracer diversity:} A major advantage of self-supervised learning is its ability to exploit large and heterogeneous pools of unlabeled data. However, several existing PET/CT pretraining studies have been conducted using relatively modest cohorts, including 752, 1,292, 507, and 717 scans in~\cite{yazdani2024automated,oh2025fratmae,huang2025c2maot,keshavarz2026self}, respectively. In addition to the limited scale, most existing approaches are developed using a single tracer, with FDG-based cohorts used in~\cite{huang2025c2maot,liu2026open,keshavarz2026self} and PSMA-based data used in~\cite{yazdani2024automated}. Although Oh \textit{et al.} incorporated data with greater tracer diversity~\cite{oh2025fratmae}, dataset size was modest (1,292) limiting diverse feature learning, and systematic evaluation across FDG and PSMA tracers were not explored. Together, the limited cohort size and  tracer diversity may constrain the ability of self-supervised pretraining to capture the full spectrum of anatomical, pathological, image acquisition related, and tracer-specific and tracer-agnostic variations encountered in clinical PET/CT imaging. As new radiotracers are being developed and used for different clinical applications, such limited training and tracer-specific models may not yield generalizable representations that are tracer-adaptable. Larger and more heterogeneous multi-institution, multi-tracer cohorts are therefore needed to develop more robust and transferable PET/CT representations.

    \item \textbf{Limited cross-modal interaction modeling:} PET and CT provide complementary information, with PET capturing molecular and physiological activity and CT providing anatomical and structural context. Features in both modalities also vary based on the anatomical region in whole-body and whether the region contains cancer or not. Clinicians use these imaging features on both modalities during interpretation, but, most existing pretraining approaches do not explicitly learn these relationships between PET and CT images. Yazdani \textit{et al.}~\cite{yazdani2024automated}, for example, followed the original SwinUNETR pretraining proxy tasks, including image rotation prediction, image volume inpainting, and contrastive coding~\cite{hatamizadeh2021swin}, while Keshavarz \textit{et al.} adopted a conventional masked-reconstruction strategy without explicitly modeling modality-specific interactions. More recent approaches~\cite{oh2025fratmae,huang2025c2maot,liu2026open} introduce modality-complementary masking, in which one modality is masked while the complementary modality remains available. Although this strategy encourages cross-modal information learning, its robustness under modality corruption, imaging noise, and distribution shifts remains insufficiently characterized.

    \item \textbf{Narrow and clinically limited downstream evaluation:} A robust pretrained representation should transfer across multiple clinically relevant tasks and remain reliable under heterogeneous and potentially degraded imaging conditions (e.g., poor image quality, artifacts, PET-CT misalignment due to patient movement, or missing modality). However, the majority of existing PET/CT self-supervised pretraining studies assess their representations primarily through lesion segmentation~\cite{yazdani2024automated,huang2025c2maot,keshavarz2026self,liu2026open}. Oh \textit{et al.} additionally evaluate staging classification~\cite{oh2025fratmae}, but lesion-level classification tasks, such as distinguishing malignant from normal patches, remain comparatively underexplored. Such region-level classification is clinically important in triage settings for flagging if a patient has cancer in a specific region or in the whole body, that may warrant more detailed investigation. Moreover, segmentation performance is commonly evaluated using voxel-level metrics such as Dice score, intersection-over-union (IoU), and Hausdorff distance. While these measures quantify spatial overlap, they do not fully capture clinically meaningful performance for staging, treatment planning, and response assessment, such as whether and how many individual lesions are correctly detected or missed, and whether specific anatomic regions contain cancer. Lesion-level sensitivity, specificity, precision, false-positive burden, and anatomical-region-specific performance therefore provide important clinically-relevant information. Last but not the least, performance robustness under likely PET/CT image quality issues, including image noise, missing modality and PET/CT misalignment has received limited attention, despite being essential for assessing model reliability under real-world acquisition variability. A comprehensive evaluation of pretrained representations should therefore span multiple downstream tasks, clinically relevant lesion-level metrics, anatomically stratified analysis, and robustness testing under realistic imaging perturbations.

\end{enumerate}



In this work, we address the above limitations of existing self-supervised pretraining approaches for whole-body PET/CT by developing and validating a novel, multimodal self-supervised learning method with context-aware joint unimodal reconstruction and cross-modal translation. Our method, Lost or Replaced Masked Image Modeling (LoRe-MIM), explicitly learns PET--CT relationships, unimodal structural consistency and anatomy from a large-scale, multi-tracer, multi-institutional pretraining cohort. We evaluate the learned representations through comprehensive downstream evaluation across segmentation and classification tasks. In addition to conventional predictive performance, we evaluate clinically relevant lesion-level performance and robustness under clinically-relevant imaging perturbations. Our main contributions are summarized as follows:

\begin{enumerate}
    \item \textbf{Large and heterogeneous PET/CT cohort:} We combine publicly available datasets, including AutoPET-III ($N=1,611$), SPADE ($N=1,126$), ViMED ($N=1,995$), and Deep-PSMA ($N=200$), with an institutional cohort from Dartmouth Hitchcock Medical Center (DHMC; $N=1,599$), yielding a total of 6,531 whole-body PET/CT scans, of which 5,910 are used for self-supervised pretraining and 621 for evaluation. The cohort spans multiple cancer types, including melanoma, lung cancer, lymphoma, and prostate cancer, as well as multiple PET tracers, including FDG and PSMA. This large-scale collection captures substantial real-world heterogeneity in patient populations, disease distributions, scanners, institutions, acquisition settings, and tracer characteristics, providing a diverse foundation for learning transferable PET/CT representations.

    \item \textbf{Lost or Replaced Masked Image Modeling (LoRe-MIM):} We introduce a novel multimodal masked image modeling strategy called {Lost or Replaced Masked Image Modelling (LoRe-MIM)} that mimics a situation of solving two jumbled-up puzzles, with some puzzle pieces that are `lost' and some pieces are `replaced' with a corresponding piece from the other puzzle. Rather than relying solely on conventional masking~\cite{keshavarz2026self} or keeping one modality fully intact while masking the other~\cite{huang2025c2maot,liu2026open}, our approach combines standard zero-mean masking with cross-modal replacement of spatially corresponding PET and CT blocks. During cross-modal replacement, a corrupted block from one modality is replaced by the co-located content from the complementary modality while the reconstruction target remains the original block. This hybrid objective encourages the model to jointly learn intra-modality contextual structure through the zero-mean masking, and bidirectional PET$\leftrightarrow$CT correspondences through the context-aware translation within a unified self-supervised framework.
    
    \item \textbf{Extensive clinically relevant downstream evaluation:} We evaluate the learned representations through a variety of clinically-relevant downstream tasks including (a) whole-body multimodal image reconstruction, (b) patch-level classification into cancer vs non-cancer, (c) robustness evaluation with patch-based classification, (d) whole-body lesion segmentation, and (e) label-efficient segmentation training.  We evaluate whole body reconstruction performance for both PET and CT images to assess how well the model learns modality-specific anatomic features as well as PET-CT relationships. For patch-based classification tasks, we assess performance with clean as well as degraded imaging conditions to evaluate robustness. For lesion segmentation, beyond conventional voxel-level segmentation metrics, we report clinically relevant lesion-level performance and stratified results across anatomical regions, PET tracers, different cancer types, and different multi-institutional datasets.

\end{enumerate}

Our experiments demonstrate that pretraining with large-scale, multi-tracer, pan-cancer  heterogeneous PET/CT datasets improves lesion segmentation, label-efficient training, classification, robustness to image quality degradation and reconstruction. Our proposed LoRE-MIM improves performance across all downstream tasks over conventional MAE-style and cross-modal masking baselines, providing generalizable, tracer-adaptable representations that can be used for different PET/CT tasks, cancers and PET tracers. 

\section{Data and Methods}

\subsection{Datasets}
\label{sec:datasets}

We utilize five retrospective, de-identified PET/CT datasets, as summarized in Table~\ref{tab:datasets}. The combined cohort from AutoPET-III, DHMC, SPADE, and ViMED comprises 6,331 scans, partitioned into 4,823 training, 1,087 validation, and 421 held-out test scans. An additional 200 scans from Deep-PSMA are reserved as an out-of-domain test set for whole-body lesion segmentation.

\subsubsection{Public datasets}
AutoPET-III~\cite{gatidis2022fdg,ingrisch2024autopet} contains 1,611
whole-body PET/CT studies spanning FDG and PSMA tracers, multiple cancer
types, and lesion-negative controls. Its FDG cohort includes melanoma, lung
cancer, and lymphoma, whereas its PSMA cohort primarily represents prostate
cancer. Expert-verified lesion annotations are available for downstream
segmentation.

Deep-PSMA~\cite{jackson2025deeppsma} contains 100 patients with metastatic
prostate cancer, each imaged using both FDG- and PSMA-PET/CT before
$^{177}$Lu-PSMA therapy, yielding 200 expert-annotated scans. Deep-PSMA is
excluded from pretraining and used only as an external segmentation test set.

SPADE~\cite{eyuboglu2021weak} contains whole-body FDG-PET/CT examinations
acquired during routine staging and restaging. VI-MED
~\cite{nguyen2025vietnamese} provides additional FDG-PET/CT data with
demographic and acquisition diversity. Because VI-MED stores CT intensities
in a non-standard integer range and PET data as raw emission counts rather
than standardized uptake values, all datasets are harmonized through a common
preprocessing pipeline.

\subsubsection{Internal dataset.}
The internal DHMC cohort comprises 1,599 unlabeled PSMA-PET/CT scans from
retrospective patients with or without prostate cancer treated at Dartmouth
Hitchcock Medical Center in Lebanon, New Hampshire. The study was approved by
the Dartmouth College and Dartmouth Hitchcock Medical Center Institutional
Review Boards, with informed consent waived because only previously acquired,
de-identified data were used.

\subsection{Data allocation}
Our pretraining uses the training and validation splits of AutoPET-III,
DHMC, SPADE, and VI-MED. Reconstruction is evaluated without labels on the
held-out AutoPET-III and DHMC test sets. Downstream segmentation is fine-tuned
using only the AutoPET-III training and validation sets and evaluated on the
held-out AutoPET-III and unseen Deep-PSMA test sets. Split assignment is
patient-disjoint whenever repeated scans are available, preventing scans from
the same patient from appearing in multiple splits.

\subsection{Preprocessing and patch sampling}

The model operates on paired, spatially aligned PET/CT volumes that are
resampled to a common voxel spacing, cropped to the body region, and
independently z-score normalized during preprocessing, following a strategy
similar to Liu \textit{et al.}~\cite{liu2026open}. During pretraining, one
spatially corresponding patch of size
$96\times128\times128$ voxels is uniformly sampled from each PET/CT pair.
A new crop is sampled whenever a volume is accessed, exposing the model to
different anatomical regions across epochs. If a volume is smaller than the
required patch size along any dimension, reflect padding is applied before
cropping. The resulting input is
$\mathbf{X}\in\mathbb{R}^{2\times96\times128\times128}$, containing the
aligned CT and PET patches. 
\begin{table*}[htbp]
\small
\centering
\caption{Summary of datasets used. The FDG/PSMA row marks tracer availability (\cmark\ present, \xmark\ absent). Lu=Lung, Ly=Lymphoma, Me=Melanoma, Pr=Prostate, Neg=Negative.}
\label{tab:datasets}
\begin{tabular}{@{}lccccc@{}}
\hline
Properties & AutoPET-III & DeepPSMA & DHMC & SPADE & VI-MED \\

\hline
Disease dist.
 & {\begin{tabular}{c} Me/Lu/Ly/Pr/Neg \\ 188/168/145/537/573 \end{tabular}}
 & Pr  & Pr & -- & -- \\
\hline
FDG/PSMA              & \cmark/\cmark & \cmark/\cmark & \xmark/\cmark & \cmark/\xmark & \cmark/\xmark \\

\# lesions (per case) & 17.47$\pm$46.12 & 69.99$\pm$60.04 & 10.13$\pm$16.27 & -- & -- \\
\# slices (per vol)   & 135--963 & 195--1261 & 303--380 & 103-307 & 207-551 \\
Pretraining & Yes & No & Yes & Yes & Yes\\
Segmentation & Yes & Yes & No & No & No \\
Pretraining-train/val/test scans        & 1043/247/321 & 0/0/0 & 1350/149/100 & 1013/113/0 & 1417/578/0 \\
Segmentation-train/val/test        & 1043/247/321 & 0/0/200 & 0/0/0 & 0/0/0 & 0/0/0 \\
\hline
\end{tabular}
\end{table*}

\subsection{Method}
\label{sec:method}

\subsubsection{LoRE-MIM: Lost or Replaced Masked Image Modeling for PET/CT self-supervised pretraining}

In this work, we aim to learn latent representations from whole-body PET/CT scans through self-supervised pretraining of an encoder–decoder architecture. Specifically, we adopt the Masked Image Modeling (MIM) paradigm for 3D medical imaging, in which paired and spatially aligned 3D patches are extracted from whole-body PET/CT scans and perturbed using a masking strategy. The model is then trained to reconstruct the original 3D patches from their perturbed counterparts. This reconstruction serves as a pretext task, enabling the model to learn informative representations that can subsequently be transferred to downstream tasks, such as lesion segmentation and classification. Mathematically, given a unlabelled dataset \(\mathcal{D}_\mathcal{U} = \{(X_{\mathrm{PET}}^{(i)}, X_{\mathrm{CT}}^{(i)})\}_{i=1}^{N}\) comprising \(N\) paired 3D PET and CT patches, where \(X_{\mathrm{PET}}^{(i)}, X_{\mathrm{CT}}^{(i)} \in \mathbb{R}^{D \times W \times H}\), with \(D\), \(W\), and \(H\) denoting the depth, width, and height of each patch, respectively, we aim to pretrain a model \(f_{\theta}\), parameterized by \(\theta\), to jointly reconstruct the original PET and CT patches from their perturbed counterparts, \(\tilde{X}_{\mathrm{PET}}^{(i)}\) and \(\tilde{X}_{\mathrm{CT}}^{(i)}\), as follows:

\begin{equation}\label{eq:recon_model}
(\hat{X}_{\mathrm{PET}}^{(i)},\hat{X}_{\mathrm{CT}}^{(i)})
=
f_{\theta}\left(
\tilde{X}_{\mathrm{PET}}^{(i)},
\tilde{X}_{\mathrm{CT}}^{(i)}
\right)
\end{equation}

where \(\hat{X}_{\mathrm{PET}}^{(i)}\) and \(\hat{X}_{\mathrm{CT}}^{(i)}\) denote the reconstructed 3D PET and CT patches, respectively. The model parameters are optimized by minimizing the reconstruction loss between the original and reconstructed patches:

\begin{equation}\label{eq:recon_basic}
\theta^{*}
=
\underset{\theta}{\arg\min}\;
\frac{1}{N}\sum_{i=1}^{N}
\mathcal{L}_{\mathrm{rec}}
\left(
(\hat{X}_{\mathrm{PET}}^{(i)},\hat{X}_{\mathrm{CT}}^{(i)}),
(X_{\mathrm{PET}}^{(i)},X_{\mathrm{CT}}^{(i)})
\right)
\end{equation}

Here, \(\mathcal{L}_{\mathrm{rec}}\) denotes the reconstruction loss function, which quantifies the discrepancy between the original and reconstructed 3D PET/CT patches.

\label{sec:cross_modal_pretraining}

Given a pair of spatially aligned PET and CT patches, \(X_{\mathrm{PET}},X_{\mathrm{CT}}\in\mathbb{R}^{D\times H\times W}\), we partition each patch into \(B\) non-overlapping spatial blocks and represent the patches as vectors of their constituent blocks:
\begin{equation}
 \begin{aligned}
\mathbf{X}_{\mathrm{PET}}
&=
\left[
\mathbf{x}_{\mathrm{PET}}^{(1)},
\mathbf{x}_{\mathrm{PET}}^{(2)},
\ldots,
\mathbf{x}_{\mathrm{PET}}^{(B)}
\right],\\[6pt]
\mathbf{X}_{\mathrm{CT}}
&=
\left[
\mathbf{x}_{\mathrm{CT}}^{(1)},
\mathbf{x}_{\mathrm{CT}}^{(2)},
\ldots,
\mathbf{x}_{\mathrm{CT}}^{(B)}
\right],
\end{aligned}   
\end{equation}

where \(\mathbf{x}_{\mathrm{PET}}^{(b)}\) and \(\mathbf{x}_{\mathrm{CT}}^{(b)}\) denote the \(b\)-th spatially corresponding blocks from PET and CT, respectively, and \(B\) represents the total number of blocks within each patch.
If each block has spatial dimensions \(d\times h\times w\), then
\begin{equation}
    B=\frac{DHW}{dhw}.
\end{equation}

Next, we define two binary masking vectors, \(\mathbf{M}_{\mathrm{PET}}\) and \(\mathbf{M}_{\mathrm{CT}}\), for PET and CT, respectively, as follows:
\begin{equation}
\begin{aligned}
\mathbf{M}_{\mathrm{PET}}
&=
\left[
m_{\mathrm{PET}}^{(1)},
m_{\mathrm{PET}}^{(2)},
\ldots,
m_{\mathrm{PET}}^{(B)}
\right],\\[6pt]
\mathbf{M}_{\mathrm{CT}}
&=
\left[
m_{\mathrm{CT}}^{(1)},
m_{\mathrm{CT}}^{(2)},
\ldots,
m_{\mathrm{CT}}^{(B)}
\right].
\end{aligned}
\end{equation}

Here, \(m_k^{(b)}\in\{0,1\}\), where \(k\in\{\mathrm{PET},\mathrm{CT}\}\), indicates whether the \(b\)-th block is selected for masking (\(m_k^{(b)} = 1\)) or remains unmasked (\(m_k^{(b)}=0\)). The two masking vectors are independently sampled, with exactly \(K\) blocks selected from each modality.

We introduce two modality-specific binary random vectors, \(Z_k = [z_k^{(1)}, \ldots, z_k^{(B)}]\), where \(k \in \{\mathrm{PET}, \mathrm{CT}\}\) denotes the imaging modality and \(z_k^{(b)} \sim \operatorname{Bernoulli}(0.5)\), to independently determine whether each selected block undergoes a \textbf{Lost} or \textbf{Replaced} perturbation with equal probability. Each pair of spatially corresponding PET and CT blocks, \(\mathbf{x}_{\mathrm{PET}}^{(b)}\) and \(\mathbf{x}_{\mathrm{CT}}^{(b)}\), is perturbed according to its modality-specific masking indicator \(m_k^{(b)}\) and binary perturbation indicator \(z_k^{(b)}\), and we achieve the perturbed PET and CT blocks, $\widetilde{\mathbf{x}}_{\mathrm{PET}}^{(b)}$ and $\widetilde{\mathbf{x}}_{\mathrm{CT}}^{(b)}$ as follows:
\begin{equation}\label{eq:LoRe_eq}
\begin{aligned}
\widetilde{\mathbf{x}}_{\mathrm{CT}}^{(b)}
&=
(1-m_{\mathrm{CT}}^{(b)})\mathbf{x}_{\mathrm{CT}}^{(b)}
+
m_{\mathrm{CT}}^{(b)}(1-z_{\mathrm{CT}}^{(b)}) 
\mathbf{x}_{\mathrm{PET}}^{(b)},\\
\widetilde{\mathbf{x}}_{\mathrm{PET}}^{(b)}
&=
(1-m_{\mathrm{PET}}^{(b)})\mathbf{x}_{\mathrm{PET}}^{(b)}
+
m_{\mathrm{PET}}^{(b)}(1-z_{\mathrm{PET}}^{(b)})
\mathbf{x}_{\mathrm{CT}}^{(b)}.
\end{aligned}
\end{equation}

Here, \(b\in\{1,\ldots,B\}\). When \(m_k^{(b)}=0\), the original block remains unchanged. When \(m_K^{(b)}=1\), the block is either replaced with zeros (\(z_k^{(b)}=1\), Lost) or substituted with its spatially corresponding block from the complementary modality (\(z_k^{(b)}=0\), Replaced). 

Finally, the perturbed blocks are spatially rearranged according to their original positions to reconstruct the 3D perturbed PET and CT patches, \(\widetilde{\mathbf{X}}_{\mathrm{PET}}\) and \(\widetilde{\mathbf{X}}_{\mathrm{CT}}\), respectively, as follows:
\begin{equation}
\begin{aligned}
\widetilde{\mathbf{X}}_{\mathrm{PET}}
&=
\operatorname{Reassemble}
\left(
\widetilde{\mathbf{x}}_{\mathrm{PET}}^{(1)},
\widetilde{\mathbf{x}}_{\mathrm{PET}}^{(2)},
\ldots,
\widetilde{\mathbf{x}}_{\mathrm{PET}}^{(B)}
\right),\\[6pt]
\widetilde{\mathbf{X}}_{\mathrm{CT}}
&=
\operatorname{Reassemble}
\left(
\widetilde{\mathbf{x}}_{\mathrm{CT}}^{(1)},
\widetilde{\mathbf{x}}_{\mathrm{CT}}^{(2)},
\ldots,
\widetilde{\mathbf{x}}_{\mathrm{CT}}^{(B)}
\right).
\end{aligned}
\end{equation}

Here, \(\operatorname{Reassemble}(\cdot)\) denotes the spatial reconstruction operation that restores the original block arrangement, yielding perturbed patches of dimensions \(D\times H\times W\).

\begin{figure*}[htbp]
    \centering
    \includegraphics[width=1\textwidth]{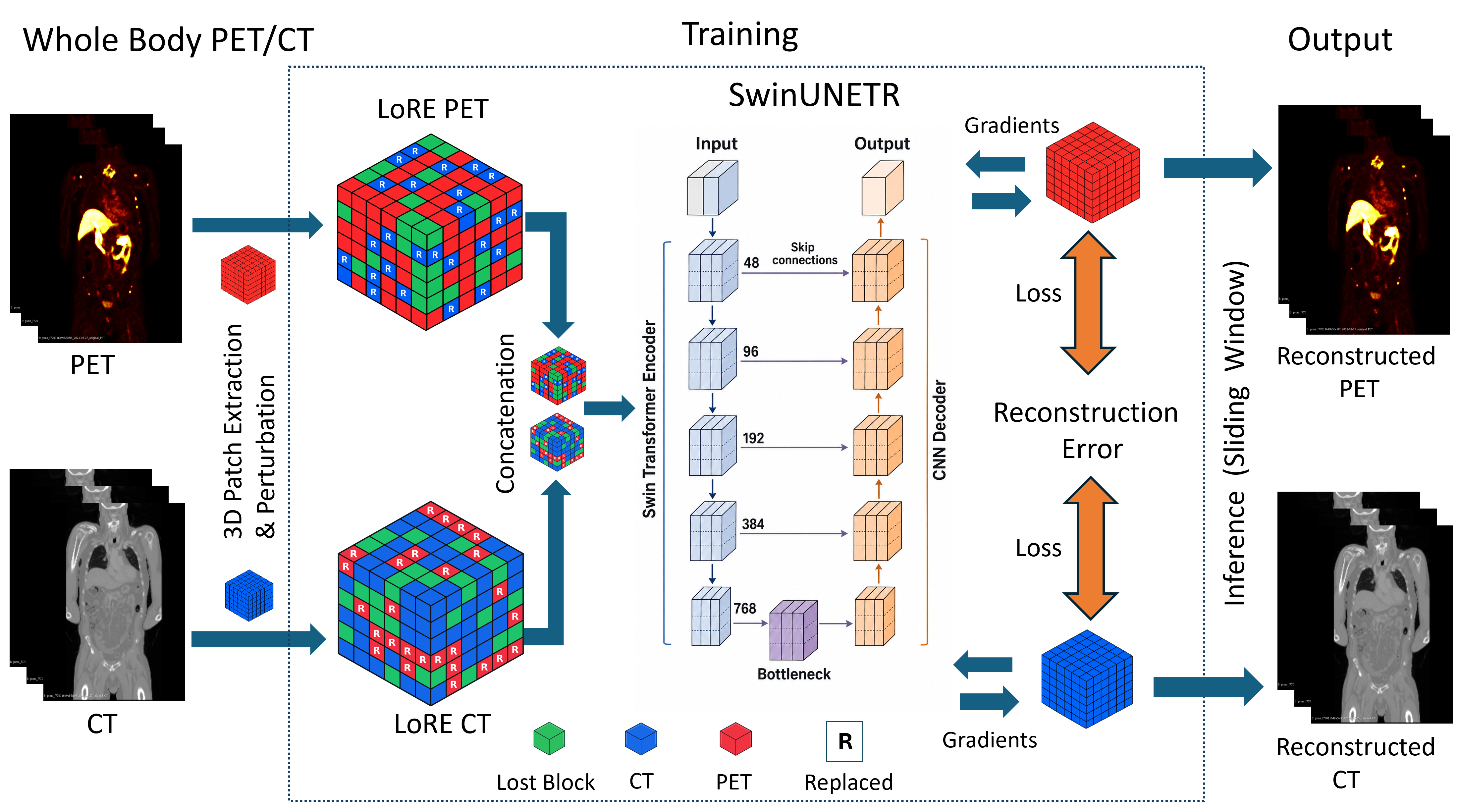}
    \caption{Overview of the proposed masking approach.}
    \label{fig:loremim_over}
\end{figure*}

An overview of our proposed $\textbf{LoRe-MIM}$ is provided in Figure \ref{fig:loremim_over}. 

\subsubsection{Reconstruction Objective}
\label{sec:reconstruction_objective}

As outlined in Eq.~\ref{eq:recon_model}, the perturbed 3D PET/CT
patches, $\widetilde{\mathbf{X}}_{\mathrm{PET}}^{(i)}$ and
$\widetilde{\mathbf{X}}_{\mathrm{CT}}^{(i)}$, are fed into the
model $f_{\theta}$ to obtain the corresponding reconstructed
patches, $\widehat{\mathbf{X}}_{\mathrm{PET}}^{(i)}$ and
$\widehat{\mathbf{X}}_{\mathrm{CT}}^{(i)}$. Following the
reconstruction objective defined in Eq.~\ref{eq:recon_basic},
we optimize the model parameters $\theta$ by minimizing the
discrepancy between the original and reconstructed 3D PET/CT
patches. We adopt the standard least squares $L_2$ loss as the
reconstruction objective. However, rather than assigning equal importance to all voxels
within a 3D patch, we introduce a block-wise weighting strategy
that prioritizes cross-modally replaced blocks over lost blocks and unchanged blocks. Specifically,
we assign a unit weight to the reconstruction errors of
replaced blocks, while downweighting those of lost and
unchanged blocks by a factor $\lambda$, where
$0 \leq \lambda < 1$. This strategy encourages the model to
emphasize cross-modal information during reconstruction while
retaining a reduced contribution from the remaining blocks.
Accordingly, the block-wise weighted reconstruction objective
is formulated as
\begin{equation}\label{eq:pretrain_loss}
\begin{aligned}
\theta^{*}
=
\underset{\theta}{\arg\min}\;
\frac{1}{NB}
\sum_{i=1}^{N}
\sum_{k\in\{\mathrm{PET},\mathrm{CT}\}}
\Bigg[
&\underbrace{
\lambda
\sum_{\substack{
b:\,m_k^{(i,b)}=0\\
\text{or }(m_k^{(i,b)}=1,\;z_k^{(i,b)}=1)
}}
\mathcal{L}_{k}^{(i,b)}
}_{\text{Lost or unchanged}}
\\[12pt]
&+
\underbrace{
\sum_{\substack{
b:\,m_k^{(i,b)}=1\\
z_k^{(i,b)}=0
}}
\mathcal{L}_{k}^{(i,b)}
}_{\text{Replaced}}
\Bigg].
\end{aligned}
\end{equation}
where $N$ and $B$ denote the number of paired 3D PET/CT
patches and the number of blocks within each patch,
respectively, and $k\in\{\mathrm{PET},\mathrm{CT}\}$ denotes
the imaging modality. Here, $m_k^{(i,b)}$ represents the
binary masking indicator, with $m_k^{(i,b)}=1$ indicating a
selected block, while $z_k^{(i,b)}$ determines whether the
selected block undergoes a Lost ($z_k^{(i,b)}=0$) or
Replaced ($z_k^{(i,b)}=1$) perturbation. The reconstruction
error for each block is defined as
\begin{equation}
\mathcal{L}_{k}^{(i,b)}
=
\frac{1}{|\Omega_b|}
\left\|
\widehat{\mathbf{x}}_{k}^{(i,b)}
-
\mathbf{x}_{k}^{(i,b)}
\right\|_2^2,
\label{eq:block_reconstruction_error}
\end{equation}
where $\mathbf{x}_{k}^{(i,b)}$ and
$\widehat{\mathbf{x}}_{k}^{(i,b)}$ represent the original
and reconstructed blocks, respectively, and $|\Omega_b|$
denotes the number of voxels within each block.

\subsection{Evaluation of LoRE-MIM Pretraining}
We conduct several experiments to test if our self-supervised pretraining method enables learning robust features for improving whole-body reconstruction, distinguishing cancereous tissue from normal tissue via classification and lesion segmentation tasks, is robust to noise, missing modality, and PET/CT misalignment, and enables label-efficient training. We also test whether our method is generalizable across tracers and multi-institutional data.

\subsubsection{Whole-body multimodal image reconstruction}
We first evaluate the reconstruction capability of our self-supervised 3D patch-based pretraining model on whole-body multimodal PET/CT images. We employ a sliding-window inference strategy with a patch size of $96 \times 128 \times 128$ voxels to reconstruct the entire volume. Overlapping reconstructed patches are subsequently aggregated using Gaussian-weighted averaging to minimize boundary artifacts and ensure spatial continuity in the reconstructed whole-body PET/CT images. 

\paragraph{Reconstruction Evaluation Method:}
The reconstructed whole-body volumes are evaluated using three metrics: Mean Absolute Error (MAE), Structural Similarity Index (SSIM), and Peak Signal-to-Noise Ratio (PSNR). MAE quantifies the average absolute voxel-wise difference between the original and reconstructed scans. SSIM measures the structural similarity between the original and reconstructed 3D volumes, while PSNR assesses reconstruction fidelity relative to the maximum signal intensity. Overall, lower MAE and higher SSIM and PSNR values indicate better reconstruction performance. 

\subsubsection{Patch-level classification into cancer vs non-cancer}

We evaluate if the learned representations can classify 3D patches into cancer vs non-cancer, and if the classification performance varies based on the amount of cancer present in each patch. We extract features from different stages of the pretrained SwinUNETR model, including the encoder output, the outputs of the five decoder stages, and feed them through a logistic regression classifier. We investigate the following configurations for this classification task

\begin{enumerate}
        \item \textbf{Feature Aggregation:} We investigate two approaches for feature aggregation: flattening and global average pooling (GAP) to obtain a feature vector for each lesion.
        
        \item \textbf{Dimensionality Reduction:} To reduce the dimensionality of the extracted and/or aggregated feature vectors, we apply Principal Component Analysis (PCA) and  kernel principal component analysis (KPCA) with a radial basis function (RBF) kernel. We compare reduced feature representations with 128 and 384 components.
    
\end{enumerate}

\paragraph{Classification Evaluation Method:} To evaluate binary classification performance of cancer vs. non-cancer, we construct the receiver operating characteristic (ROC) curve and compute the area under the ROC curve (AUC-ROC). 

\subsubsection{Robustness evaluation with patch-based classification}\label{sec:robust_classification}
We evaluate the robustness of the learned representations for patch classification using synthetically perturbed 3D PET/CT patches under three clinically relevant scenarios: (1) noisy images, (2) spatial misalignment between PET and CT modalities due to registration errors, and (3) missing modality information, where either PET or CT is unavailable. 
\begin{itemize}
    \item \textbf{Robustness to Noisy Images:} We independently introduce zero-mean Gaussian noise into PET and CT patches of testing cohort to evaluate the robustness of the learned representations to image corruption. These perturbations provide a controlled approximation of image quality degradation that may arise from variations in image acquisition and reconstruction. Specifically, Gaussian noise is independently added to the PET and CT voxels, with standard deviations set to 0.1 times the corresponding modality's intensity standard deviation computed from the clean training data. The resulting corrupted intensities are clipped to the intensity range of the corresponding modality in the clean training data.   
    \item \textbf{Robustness to PET/CT misalignment:} Spatial misalignment between PET and CT may arise from patient movement between acquisitions~\cite{zaidi2007optimisation}, differences in respiratory states~\cite{callahan2011clinical}, and artifacts introduced by CT-based attenuation correction in the presence of respiratory motion~\cite{osman2003respiratory}. We simulate such spatial misalignment between PET and CT modalities by applying random geometric transformations to the CT channel of each test patch while keeping the PET channel unchanged. Specifically, we independently sample in-plane rotations about the $z$-axis uniformly from $[-5^{\circ}, 5^{\circ}]$ and translations along the $x$- and $y$-axes uniformly from $[-2, 2]$ voxels, while maintaining zero translation along the $z$-axis. The transformed CT patches are resampled using trilinear interpolation with zero padding. These perturbations simulate registration errors between PET and CT modalities while preserving both modalities.
    \item \textbf{Robustness to Missing Modality:} We simulate scenarios in which either the PET or CT modality is unavailable to evaluate the robustness of the learned representations to missing modality information. Specifically, we consider two scenarios: (1) Missing-PET, where all PET voxel intensities are set to zero while the CT channel remains unchanged, and (2) Missing-CT, where all CT voxel intensities are set to zero while the PET channel remains unchanged.
\end{itemize}

\paragraph{Evaluation Method:} Similar to evaluation with clean dataset, we evaluate ROC-AUC for evaluating robustness of the binary classification task. 
\subsubsection{Whole-body Lesion Segmentation}\label{sec:seg_finetuning}
 We evaluate the effectiveness of the learned representations for whole-body lesion segmentation by finetuning the pretrained model using a labeled dataset
$\mathcal{D}_{\mathcal{L}} =
\{(X_{\mathrm{PET}}^{(i)}, X_{\mathrm{CT}}^{(i)}, Y^{(i)})\}_{i=1}^{N_{\mathcal{L}}}$,
comprising $N_{\mathcal{L}}$ paired 3D PET/CT scans and their corresponding voxel-wise binary lesion segmentation masks $Y^{(i)}$. We denote the parameters of the pretrained model by $\theta$, which comprises three mutually disjoint subsets: the backbone parameters $\theta_{\mathrm{backbone}}$, encoder parameters $\theta_{\mathrm{enc}}$, and decoder parameters $\theta_{\mathrm{dec}}$, such that
$\theta = \theta_{\mathrm{backbone}} \cup \theta_{\mathrm{enc}} \cup \theta_{\mathrm{dec}}$. During finetuning, we partition $\theta$ into trainable parameters $\theta_{\mathrm{FT}}$, which are initialized from their corresponding pretrained weights and subsequently optimized using the labeled dataset, and frozen parameters $\theta_{\mathrm{frozen}}$, which remain fixed throughout finetuning. Following Rokuss \textit{et al.}~\cite{rokuss2024fdg}, we optimize the trainable parameters using an equally weighted combination of Dice loss and cross-entropy loss to encourage accurate lesion delineation and voxel-wise classification.

To investigate the contribution of self-supervised pretraining and the effects of adapting different model components to downstream lesion segmentation, we compare training from scratch against three finetuning strategies:

\begin{enumerate}
    \item \textbf{Training from Scratch (Scratch):}
    We randomly initialize and optimize all model parameters $\theta$ using the labeled dataset, without leveraging pretrained weights.

    \item \textbf{Full Finetuning (Full FT):}
    We initialize all model parameters from the pretrained weights and optimize the entire model, with
    $\theta_{\mathrm{FT}} = \theta$
    and
    $\theta_{\mathrm{frozen}} = \varnothing$.

    \item \textbf{Encoder and Decoder Finetuning (Enc. + Dec. FT):}
    We optimize the pretrained encoder and decoder parameters while keeping the backbone frozen, with
    $\theta_{\mathrm{FT}} = \theta_{\mathrm{enc}} \cup \theta_{\mathrm{dec}}$
    and
    $\theta_{\mathrm{frozen}} = \theta_{\mathrm{backbone}}$.

    \item \textbf{Decoder Finetuning (Dec. FT):}
    We optimize only the pretrained decoder parameters while keeping both the backbone and encoder frozen, with
    $\theta_{\mathrm{FT}} = \theta_{\mathrm{dec}}$
    and
    $\theta_{\mathrm{frozen}} = \theta_{\mathrm{backbone}} \cup \theta_{\mathrm{enc}}$.
\end{enumerate}

\paragraph{Lesion Segmentation Evaluation Method:} We adopt voxel-level, lesion-level, and anatomical region-level evaluation metrics as below: 

\begin{enumerate}
   
\item \textbf{Voxel-level Evaluation:} Following Rokuss \textit{et al.}~\cite{rokuss2024fdg}, we use the Dice similarity coefficient, false-positive volume (FPVol), and false-negative volume (FNVol) as voxel-level evaluation metrics. Dice quantifies the spatial overlap between the predicted and ground truth segmentations. FPVol measures the total volume of tissue incorrectly predicted as lesions, whereas FNVol measures the total volume of ground truth lesion tissue missed by the model. We report these metrics only for diseased cases.

\item \textbf{Lesion-level Evaluation:} We identify individual lesions in the predicted and ground truth segmentation masks using morphological processing and connected component analysis. We then aggregate lesion-level counts across all test patients and compute sensitivity, defined as the proportion of ground truth lesions detected, and positive predictive value (PPV), defined as the proportion of predicted lesions that match ground truth lesions.

\item \textbf{Anatomical Region-level Evaluation:}
Treatment decisions are often influenced by the presence of metastases in specific anatomical sites, such as the liver or bones. We therefore evaluate segmentation performance in four anatomical regions: the liver, lungs, bones, and high-risk bones. We use TotalSegmentator~\cite{wasserthal2023totalsegmentator} to obtain anatomical masks and compute region-level Dice, FPVol, and FNVol using the voxel-level definitions restricted to each region. We also compute region-level sensitivity, defined as the proportion of ground truth lesion-containing regions in which at least one lesion is correctly detected, and specificity, defined as the proportion of ground truth lesion-free regions in which no lesion is predicted.

\subsubsection{Label-Efficient Segmentation Training}

Obtaining voxel-wise annotations for whole-body lesion segmentation is a time-consuming and resource-intensive process that requires substantial clinical expertise. As discussed in Sect.~\ref{sec:intro}, the growing volume of whole-body PET/CT examinations further widens the gap between imaging demand and the availability of expert radiologists to generate or verify such annotations. Therefore, we evaluate the effectiveness of self-supervised pretraining under label-efficient finetuning settings by utilizing varying proportions (1\%, 5\%, 10\%, 20\%, 50\%, and 100\%) of the labeled training data for whole-body lesion segmentation.

\paragraph{Evaluation Method:} We evaluate segmentation performance using Dice and report the results across all evaluated percentages of labeled training data.

\end{enumerate}







\noindent

\subsection{Implementation Details}

\label{sec:pretraining_implementation}
We adopt the 3D SwinUNETR~\cite{hatamizadeh2021swin} architecture for our pretraining model using its MONAI~\cite{cardoso2022monai} implementation with a feature size of 48. The Swin Transformer encoder comprises four hierarchical stages with depths of (2, 2, 6, 2) and attention heads of (3, 6, 12, 24), respectively. The network operates on 3D volumetric PET/CT inputs, with the output channels configured to reconstruct both PET and CT images during pretraining and to generate binary foreground and background segmentation logits during downstream fine-tuning. implementation.  
\begin{enumerate}
    \item \textbf{SwinUNETR pretraining:} We set the weighting parameter $\lambda=0.2$ in Eqn.~\ref{eq:pretrain_loss} to assign a lower contribution to unperturbed or zero-masked blocks than to cross-modally perturbed blocks in the pretraining loss. We train the complete network for 200 epochs using AdamW with a batch size of
3, a maximum learning rate of $10^{-4}$, and weight decay of $10^{-2}$.
Training uses automatic mixed precision and gradient checkpointing. The
learning rate is linearly increased from $10^{-6}$ to $10^{-4}$ during the
first five epochs and then decayed to $10^{-6}$ using cosine annealing over
the remaining epochs. Validation is performed after every epoch, and the
checkpoint with the lowest total validation reconstruction loss is retained. 
for downstream evaluation. 
    \item \textbf{Classification models:} We train logistic regression classifiers with our extracted features with balanced class weights, and a maximum of 2,000 optimization iterations. The inverse regularization strength is selected from $C \in \{10^{-3},10^{-2},10^{-1},1,10,10^{2},10^{3}\}$ by maximizing ROC AUC on the clean validation split. Each candidate classifier is fitted only on training features.

    \item \textbf{Segmentation models:}  We employ the nnU-Net framework adapted by Rokuss \textit{et al.}~\cite{rokuss2024fdg} to finetune our \textbf{LoRe-MIM}-pretrained SwinUNETR. We train the model for 600 epochs using stochastic gradient descent (SGD) with Nesterov momentum and a polynomial learning rate decay schedule, with an initial learning rate of 0.001 and a weight decay of $3\times10^{-5}$. We optimize the model using an equally weighted combination of Dice and cross-entropy losses and select the checkpoint with the lowest validation loss. For each fine-tuning variant described in Sect.~\ref{sec:seg_finetuning}, we initialize the frozen network components with their corresponding pretrained weights from the pretraining phase, while the remaining trainable parameters are randomly initialized and optimized during fine-tuning.
\end{enumerate}

\section{Ablation Studies for Pretraining Approaches}
\subsection{Masking Strategies}\label{sec:abla_masking}
We conduct an ablation study examining three special cases of the proposed Lost-or-Replaced (LoRe) masking framework to investigate the contributions of different masking strategies to self-supervised representation learning. A Lost perturbation sets the selected block to zero, whereas a Replaced perturbation substitutes the spatially corresponding block from the other modality. Based on this formulation, we consider three masking variants:

\begin{enumerate}
    \item \textbf{Lost-One MIM (LO-MIM):} Following MUST-PET~\cite{biswas2026must}, we apply zero masking to one modality while leaving the other unchanged. For PET masking, we sample \(m_{\mathrm{PET}}^{(b)}\sim\operatorname{Bernoulli}(0.5)\) and set \(z_{\mathrm{PET}}^{(b)}=1\) and \(m_{\mathrm{CT}}^{(b)}=0\). Conversely, for CT masking, we sample \(m_{\mathrm{CT}}^{(b)}\sim\operatorname{Bernoulli}(0.5)\) and set \(z_{\mathrm{CT}}^{(b)}=1\) and \(m_{\mathrm{PET}}^{(b)}=0\). This strategy evaluates representation learning when only one modality is perturbed.
    \item \textbf{Lost-Both MIM (LB-MIM):} We apply the same zero-masking pattern to both modalities by sampling a shared mask \(m^{(b)}\sim\operatorname{Bernoulli}(0.5)\) and setting \(m_{\mathrm{PET}}^{(b)}=m_{\mathrm{CT}}^{(b)}=m^{(b)}\) and \(z_{\mathrm{PET}}^{(b)}=z_{\mathrm{CT}}^{(b)}=1\). Thus, spatially corresponding PET and CT blocks are simultaneously set to zero, removing direct input information from both modalities at the selected locations.
    \item \textbf{Replaced-MIM (Re-MIM):} We apply only cross-modal replacement by setting \(z_{\mathrm{PET}}^{(b)}=z_{\mathrm{CT}}^{(b)}=0\) and independently sampling \(m_{\mathrm{PET}}^{(b)},m_{\mathrm{CT}}^{(b)}\sim\operatorname{Bernoulli}(0.5)\). Each selected PET block is replaced with the corresponding original CT block, and each selected CT block is replaced with the corresponding original PET block. This strategy isolates the contribution of cross-modal replacement without introducing zero masking.
\end{enumerate}

\subsection{Augmentation}\label{sec:aug}

We investigate the impact of data augmentation during self-supervised pretraining on reconstruction and patch classification (cancer vs. non-cancer) performance. We adopt the whole-body PET/CT-specific augmentation strategies employed by Rokuss \textit{et al.}~\cite{rokuss2024fdg}, incorporating both spatial and intensity-based transformations. Spatial augmentations include random rotations, scaling, and mirroring, while intensity-based augmentations include Gaussian noise, Gaussian blurring, brightness scaling, contrast adjustment, and gamma correction. Additionally, we simulate reduced image resolution and spatial misalignment between PET and CT modalities to account for variations in image quality and intermodal registration.

\subsection{Variable Masking}\label{sec:var_mask}

We investigate the effect of masking variable portions of each modality in both \textbf{Re-MIM} and \textbf{LoRe-MIM} pretraining.
For each sampled training patch of $96\times128\times128$ voxels, the CT and
PET masking ratios are drawn independently with equal probability.
Randomly selected, non-overlapping blocks of $12\times16\times16$ voxels are masked according to the sampled ratio for each modality, with new ratios and mask locations sampled whenever a training patch was drawn.
Both the ablation masking strategy (Re-MIM and LoRe-MIM) retains its original replacement strategy. We evaluated variable masking alone and in combination with augmentation. Validation used no augmentation and a fixed masking ratio of $0.5$ for
each modality.

\section{Baseline Models For Comparison}

We compare our pretraining approach with reconstruction performance as well as downstream tasks such as segmentation and classification. 

\subsection{Reconstruction}

We compare our proposed \textbf{LoRe-MIM} with an off-the-shelf whole-body self-supervised model trained with FDG PET/CT, released by Liu \textit{et al.}~\cite{liu2026open}. We denote this baseline as \textbf{FDG-only MAE}. 

\subsection{Segmentation Models} 

We consider two baseline models for comparison with our finetuned segmentation models: (1) SwinUNETR trained from scratch and (2) the whole-body PET/CT-based nnU-Net architecture proposed by~\cite{rokuss2024fdg}, also trained from scratch. For fair comparison, we utilize the same preprocessed samples, same training framework released by~\cite{rokuss2024fdg} with similar augmenttaion and same number of training epochs.

\subsection{Classification performance with pretrained features}

We compare our pretrained representations with those extracted from DINOv3~\cite{simeoni2025dinov3} and ResNet50-3D~\cite{chen2019med3d}. For both baselines, PET and CT patches are independently normalized using their respective means and standard deviations (with a minimum standard deviation of $10^{-6}$) and concatenated side by side. 
For ResNet50-3D, global average pooling of the final feature map produces a 2048-dimensional feature vector. For DINOv3, each axial slice is resized to $224 \times 224$, replicated across three channels, and processed independently. The resulting slice-level features are averaged to obtain a single feature vector per patch.

\section{Experimental Results}

\subsection{Validation Set Performance for ablation studies}
We report validation set performance for the ablation studies to select the best pretraining strategy. 

\subsubsection{Different masking strategies}

We assess reconstruction and patch-based classification performance on the validation set for the ablation experiments. For reconstruction performance, we consider the validation set of AutoPET-III ($N=247$)  and the unlabeled DHMC ($N=149$) dataset.  For classification performance, we consider the validation set of AutoPET-III, with patch-based labels derived from voxel-level annotations.
Our proposed \textbf{LoRe-MIM} consistently outperforms the three alternative masking strategies, achieving the lowest MAE and the highest SSIM and PSNR for the reconstruction task, and the highest AUC for patch classification task, demonstrating the effectiveness of its learned representations across both tasks ~\ref{tab:pretraining_ablation}.




\begin{table}[!htbp]
\centering
\caption{
Ablation study of masking strategies under fixed masking.
Reconstruction performance on AutoPET ($n=247$) and DHMC ($n=149$)
validation samples, averaged equally across CT and PET and across all
396 cases, is evaluated using MAE ($\downarrow$), SSIM ($\uparrow$),
and PSNR in dB ($\uparrow$).
Classification performance on clean test data is evaluated using
AUC ($\uparrow$). Bold indicates the best result for each metric.
}
\label{tab:pretraining_ablation}


\begin{tabular}{lcccc}
\toprule
\multirow{2}{*}{\textbf{Masking Approach}}
& \multicolumn{3}{c}{\textbf{Reconstruction}}
& \textbf{Classification} \\
\cmidrule(lr){2-4}
\cmidrule(lr){5-5}
& MAE $\downarrow$ & SSIM $\uparrow$
& PSNR $\uparrow$ & AUC $\uparrow$ \\
\midrule
FDG-Only MAE~\cite{liu2026open}
& 0.2797 & 0.7288 & 32.56 & 0.6195 \\
LO-MIM
& 0.4026 & 0.6473 & 30.01 & 0.6885 \\
LB-MIM
& 0.3831 & 0.6835 & 30.14 & 0.6894 \\
Re-MIM
& 0.2803 & 0.7586 & 32.73 & 0.7010 \\

\textbf{LoRE-MIM}
& \textbf{0.2675} & \textbf{0.7762} & \textbf{33.07} & \textbf{0.7284} \\
\bottomrule
\end{tabular}
\end{table}

\subsubsection{Augmentation and Variable Masking}

Results from both reconstruction (MAE, SSIM, and PSNR) and patch-based lesion classification (AUC) indicate that neither augmentation nor variable masking consistently improves representation quality (Table~\ref{tab:reconstruction_augmentation_ablation}). Therefore, in subsequent experiments, we adopt a fixed masking strategy with a Bernoulli probability of 0.5, without data augmentation.

\begin{table}[!htbp]
\centering
\caption{Ablation study of augmentation and variable masking for CM and LoRE-MIM on AutoPET-III ($n=321$) and DHMC ($n=100$) test sets. Reconstruction metrics are evaluated over masked regions after whole-volume reconstruction, averaged equally across CT and PET and across all 421 cases. Masked blocks are replaced with Gaussian samples of standard deviation 0.1. PSNR is in dB. Classification AUC values are retained from the original experiments. Bold indicates the best result for each metric.}
\label{tab:reconstruction_augmentation_ablation}
\begin{tabular}{lcccccc}
\toprule
\multirow{2}{*}{\textbf{Model}} & \multirow{2}{*}{\textbf{Augmentation}} & \multirow{2}{*}{\makecell{\textbf{Variable}\\\textbf{Masking}}} & \multicolumn{3}{c}{\textbf{Reconstruction}} & \textbf{Classification} \\
\cmidrule(lr){4-6}
\cmidrule(lr){7-7}
& & & MAE $\downarrow$ & SSIM $\uparrow$ & PSNR $\uparrow$ & AUC $\uparrow$ \\
\midrule
\multirow{4}{*}{Re-MIM} & $\times$ & $\times$ & 0.3913 & 0.6784 & 29.86 & 0.6855 \\
 & $\checkmark$ & $\times$ & 0.4131 & 0.6513 & 29.70 & 0.5783 \\
 & $\times$ & $\checkmark$ & 0.4258 & 0.6568 & 28.82 & 0.5695 \\
 & $\checkmark$ & $\checkmark$ & 0.4989 & 0.5738 & 28.35 & 0.5525 \\
\midrule
\multirow{4}{*}{\textbf{LoRE-MIM}} & $\times$ & $\times$ &  \textbf{0.2718} & 0.7701 & \textbf{32.85} & \textbf{0.7091} \\
 & $\checkmark$ & $\times$ &0.2727 & \textbf{0.7847} & 32.46 & 0.6042 \\
 & $\times$ & $\checkmark$ & 0.2814 & 0.7559 & 32.64 & 0.6142 \\
 & $\checkmark$ & $\checkmark$ & 0.2870 & 0.7721 & 32.42 & 0.5970 \\
\bottomrule
\end{tabular}%
\end{table}

\subsection{Test Set Results for Downstream Tasks}

\subsubsection{Reconstruction and classification performance}

We first evaluate four different masking strategies in terms of reconstruction and classification performance on the AutoPET-III and DHMC test sets (Table~\ref{tab:pretraining_ablation_test}). Test set size for reconstruction evaluation is 421 scans (AutoPet-III and DHMC), while for classification is 321 scans (AutoPet-III). Consistent with our observations on the validation set (Table~\ref{tab:pretraining_ablation}), our proposed \textbf{LoRe-MIM} achieves the lowest MAE and highest SSIM, PSNR, and AUC among all four masking strategies, further demonstrating its effectiveness in learning transferable representations for both reconstruction and classification tasks.

\begin{table}[!htbp]
\centering
\caption{
Reconstruction and classification performance. Reconstruction is evaluated on the AutoPET-III ($n=321$) and DHMC ($n=100$) test sets using masked-region MAE ($\downarrow$), SSIM ($\uparrow$), and PSNR in dB ($\uparrow$), computed after whole-volume reconstruction and averaged equally across CT and PET modalities and across all 421 test cases. Classification performance is evaluated using AUC ($\uparrow$), with values retained from the original classification experiments.
}
\label{tab:pretraining_ablation_test}
\small
\renewcommand{\arraystretch}{1.3}
\setlength{\tabcolsep}{5pt}
\begin{tabular}{lcccc}
\toprule
\multirow{2}{*}{\textbf{Masking Approach}}
& \multicolumn{3}{c}{\textbf{Reconstruction}}
& \textbf{Classification} \\
\cmidrule(lr){2-4}
\cmidrule(lr){5-5}
& MAE $\downarrow$ & SSIM $\uparrow$
& PSNR $\uparrow$ & AUC $\uparrow$ \\
\midrule
FDG-Only
& 0.2812 & 0.7256 & 32.45 & 0.6782 \\
LO-MIM
& 0.4080 & 0.6432 & 29.74 & 0.6695 \\

LB-MIM
& 0.3913 & 0.6784 & 29.86 & 0.6626 \\
Re-MIM
& 0.2857 & 0.7510 & 32.51 & 0.6855 \\
\textbf{LoRE-MIM}
& \textbf{0.2718} & \textbf{0.7701} & \textbf{32.85} & \textbf{0.7091} \\
\bottomrule
\end{tabular}
\end{table}

\subsubsection{Patch-level classification into cancer vs non-cancer}

We first evaluate patch-based classification performance on the clean AutoPET-III test set using features extracted from the SwinUNETR encoder. For all comparisons, we employ logistic regression (LR) for classification using the extracted features, with dimensionality reduction applied where specified. As shown in Table~\ref{tab:classification_ablation}, encoder features aggregated using global average pooling (GAP) achieve a higher AUC with PCA using 384 components (0.7180) than with 128 components (0.6974). Furthermore, GAP generally outperforms feature flattening, highlighting the benefits of spatial feature aggregation. We also note that linear dimensionality reduction with PCA outperforms non-linear KPCA for 128 and 384 feature dimensions, suggesting that learned representations have good linear separability in cancer vs non-cancer patterns. We then investigate features extracted from different decoder stages and observe that decoder representations consistently outperform encoder representations. In particular, Dec$_4$ achieves the highest AUC (0.7790), surpassing both DINOv3 (0.7589) and ResNet50-3D (0.7120). These results highlight the effectiveness of intermediate decoder representations learned through LoRe-MIM pretraining for downstream patch-based classification.

\begin{table}[!htbp]
\centering
\caption{Evaluation of patch-based classification using
features extracted from different model stages and baseline
architectures under clean input conditions. We compare
feature aggregation and dimensionality reduction strategies
using logistic regression (LR) for classification.
Performance is evaluated using AUC ($\uparrow$).
Bold and underline indicate the best overall and
best encoder performance, respectively.}
\label{tab:classification_ablation}


\begin{tabular}{lccc}
\toprule
\makecell{\textbf{Feature Extraction}\\
\textbf{Stage / Model}}
& \makecell{\textbf{Feature}\\\textbf{Type}}
& \textbf{Reduction}
& \textbf{AUC}$\uparrow$ \\
\midrule

\multirow{6}{*}{Enc}
& \multirow{2}{*}{Flatten}
& KPCA (128) & 0.6898 \\
& & PCA (128) & 0.6565 \\

\cmidrule(lr){2-4}

& \multirow{4}{*}{GAP}
& KPCA (128) & 0.6808 \\
& & PCA (128) & 0.6974 \\
& & KPCA (384) & 0.7079 \\
& & PCA (384) & \underline{0.7180} \\

\midrule

Dec$_1$ & GAP & -- & 0.7339 \\
Dec$_2$ & GAP & -- & 0.7370 \\
Dec$_3$ & GAP & -- & 0.7656 \\
Dec$_4$ & GAP & -- & \textbf{0.7790} \\
Dec$_5$ & GAP & -- & 0.7680 \\

\midrule

ResNet50-3D & GAP & PCA (384) & 0.7120 \\
DINOv3 & GAP & -- & 0.7589 \\

\bottomrule
\end{tabular}
\end{table}

\subsubsection{Robustness evaluation with patch-based classification}

Table~\ref{tab:classification_ablation_robustness} presents the robustness evaluation of patch-based classification using features extracted from the fourth decoder stage (Dec$_4$) of our \textbf{LoRe-MIM} pretrained SwinUNETR under perturbed 3D PET/CT inputs. Our pretrained decoder features consistently outperform the ResNet50-3D and DINOv3 baselines across all four perturbation settings. Under Gaussian noise, our method achieves an AUC of 0.7584, compared with 0.7555 for DINOv3 and 0.7268 for ResNet50-3D. Similarly, under simulated PET/CT misalignment, our method achieves a higher AUC (0.7648) than DINOv3 (0.7286) and ResNet50-3D (0.7332). More notably, all methods exhibit substantial performance degradation when either imaging modality is missing, with missing PET resulting in considerably lower AUC than missing CT. Nevertheless, our method maintains an AUC of 0.6975 under missing CT and 0.5995 under missing PET, outperforming DINOv3 (0.6500 and 0.5344) and ResNet50-3D (0.6217 and 0.5348), respectively. The largest performance gap between our method and DINOv3 occurs when PET is missing (0.0651 AUC), suggesting that our learned representations are less sensitive to the absence of PET information. These findings support the potential of \textbf{LoRe-MIM} to capture complementary cross-modal representations and improve classification robustness under incomplete or corrupted PET/CT inputs.

\begin{table}[!htbp]
\centering
\caption{Robustness evaluation of patch-based classification
under four PET/CT input perturbations: Gaussian noise,
PET/CT misalignment, missing CT, and missing PET.
Performance is evaluated using AUC ($\uparrow$).
Bold indicates the best result for each perturbation.}
\label{tab:classification_ablation_robustness}


\begin{tabular}{llcccc}
\toprule

\textbf{Category}
& \textbf{Model}
& \makecell{\textbf{Gaussian}\\\textbf{Noise}}
& \textbf{Misaligned}
& \makecell{\textbf{CT}\\\textbf{Missing}}
& \makecell{\textbf{PET}\\\textbf{Missing}} \\

\midrule

\multirow{2}{*}{Baseline}
& ResNet50-3D
& 0.7268 & 0.7332 & 0.6217 & 0.5348 \\

& DINOv3
& 0.7555 & 0.7286 & 0.6500 & 0.5344 \\

\midrule

\textbf{Ours}
& \textbf{LoRe-MIM (Dec$_4$)}
& \textbf{0.7584}
& \textbf{0.7648}
& \textbf{0.6975}
& \textbf{0.5995} \\

\bottomrule
\end{tabular}%

\end{table}

\subsubsection{Whole-body lesion segmentation with test-set}

We use $321$ cases from AutoPET-III and $200$ cases from Deep-PSMA as independent test sets for whole-body lesion segmentation. These cases are strictly excluded from both the pretraining and finetuning stages, ensuring no overlap with the training or validation cohorts used in either stage. 

\begin{figure}[htbp]
	\centering
		\begin{subfigure}[b]{1\textwidth}
		\centering
	\includegraphics[width=\textwidth]{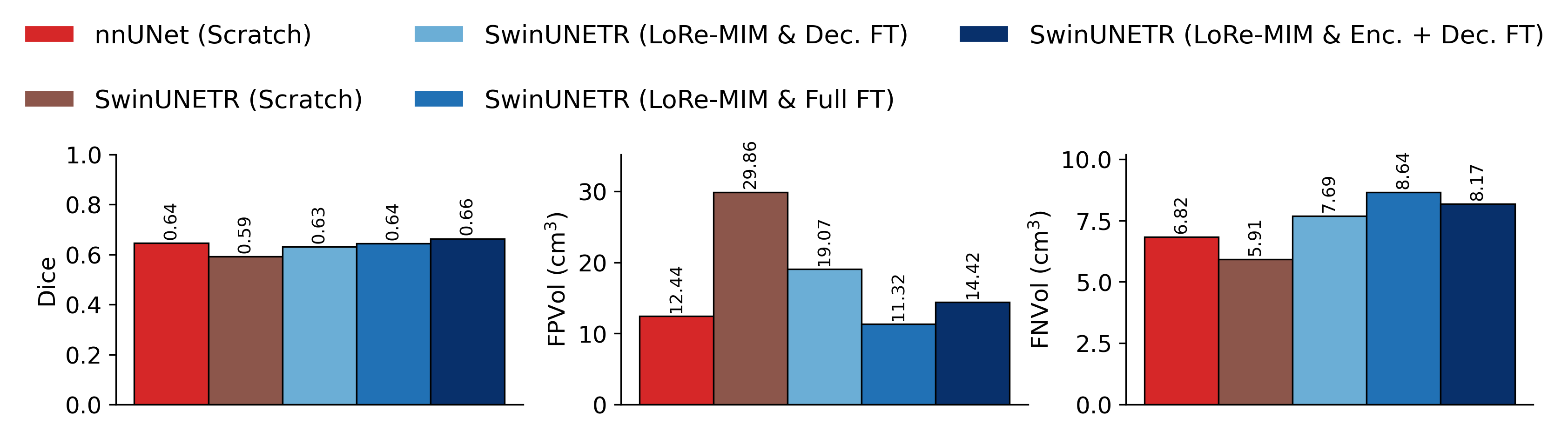}
	\caption{FDG cases ($n=199$) among a total of $321$ cases.}
    \label{figs:FDG_test_random_vs_feeds}
	\end{subfigure}
    \begin{subfigure}[b]{1\textwidth}
		\centering
	\includegraphics[width=\textwidth]{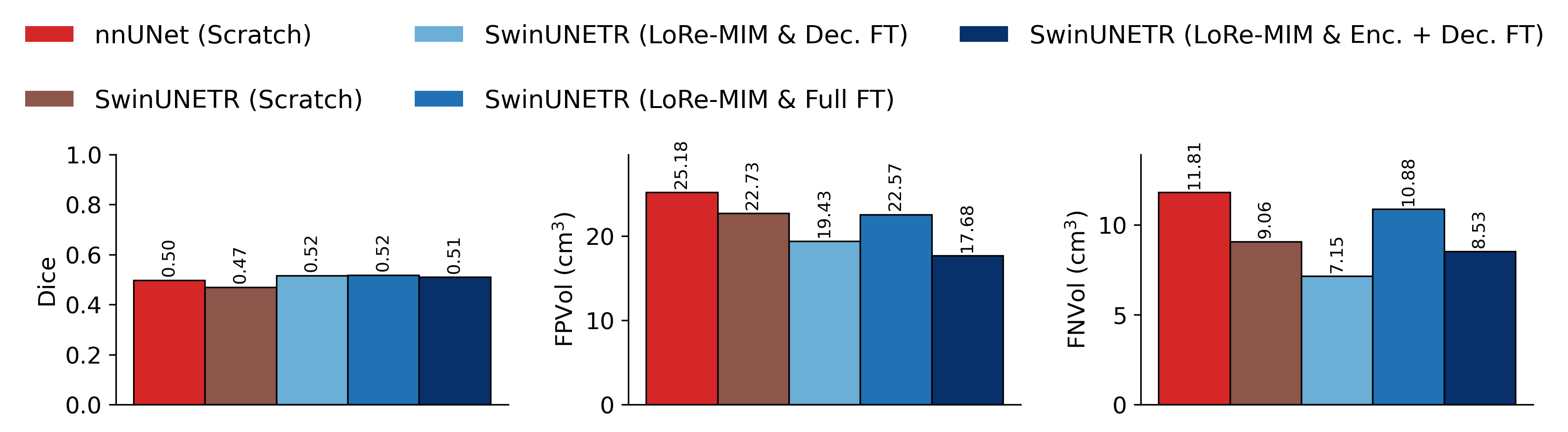}
	\caption{PSMA cases ($n=122$) among a total of $321$ cases.}
    \label{figs:PSMA_test_random_vs_feeds}
	\end{subfigure}
    \caption[example]{Tracer-specific voxel-level evaluation on the AutoPET-III test set, comparing nnU-Net and SwinUNETR trained from scratch with SwinUNETR pretrained using \textbf{LoRe-MIM} under different finetuning configurations.}  \label{fig:autopet_tracer}
\end{figure}

\begin{figure}[htbp]
	\centering
	
	\begin{subfigure}[b]{1\textwidth}
		\centering
	\includegraphics[width=\textwidth]{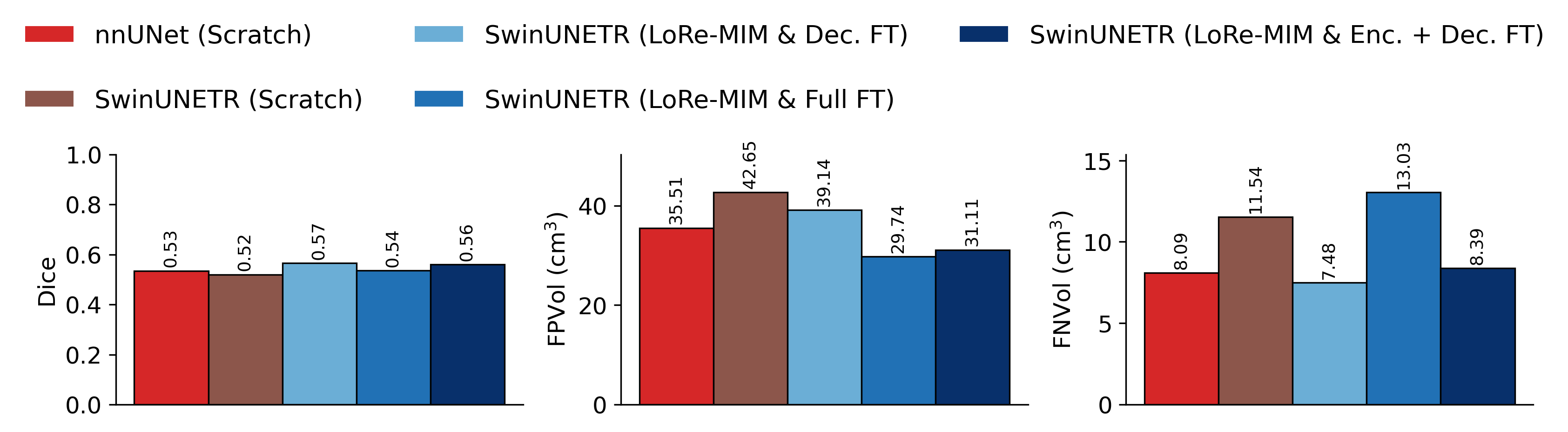}
	\caption{ FDG cases ($n=100$) among a total of $200$ cases.}
    \label{figs:FDG_test_random_vs_feeds}
	\end{subfigure}
    \begin{subfigure}[b]{1\textwidth}
		\centering
	\includegraphics[width=\textwidth]{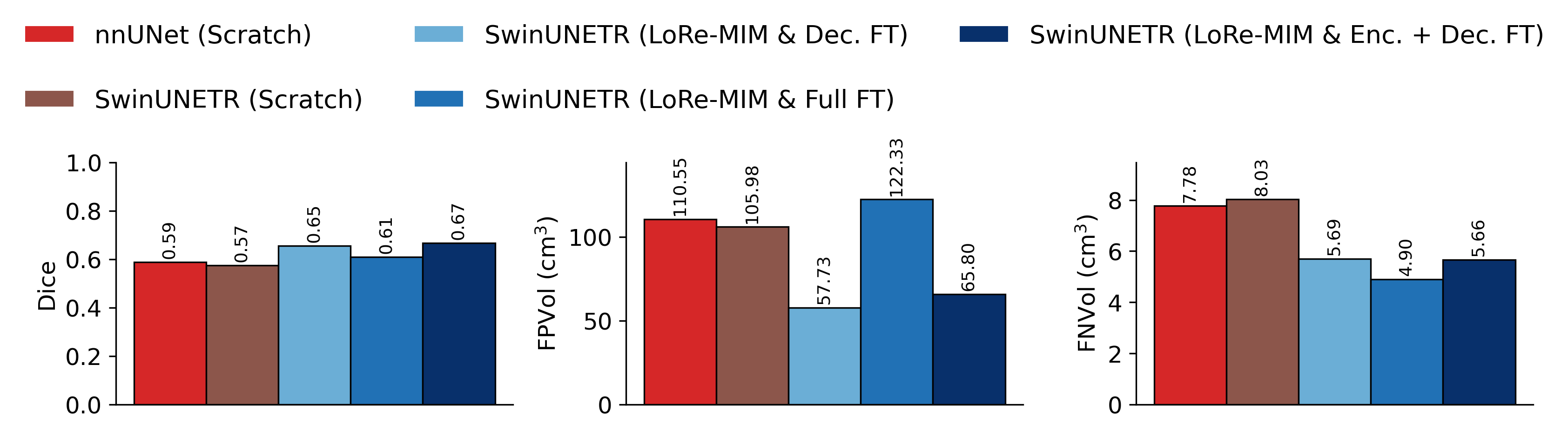}
	\caption{ PSMA cases ($n=100$) among a total of $200$ cases.}
    \label{figs:PSMA_test_random_vs_feeds}
	\end{subfigure}
    \caption[example]{Tracer-specific voxel-level evaluation on the Deep-PSMA test set, comparing nnU-Net and SwinUNETR trained from scratch with SwinUNETR pretrained using \textbf{LoRe-MIM} under different finetuning configurations.}
    \label{fig:deep_psma_tracer}
\end{figure}

\paragraph{Voxel and Lesion-level evaluation:} We compare our proposed \textbf{LoRe-MIM} pretrained SwinUNETR against SwinUNETR and nnU-Net trained from scratch, considering three finetuning configurations: Full FT, Dec. FT, and Enc.~+~Dec. FT, as described in Sect.~\ref{sec:seg_finetuning}. Among these configurations, Enc.~+~Dec. FT achieves the highest Dice scores (0.581 and 0.613) and lesion sensitivities (0.776 and 0.759) on AutoPET-III and Deep-PSMA, respectively. Compared with SwinUNETR trained from scratch, these correspond to relative Dice improvements of 10.5\% and 12.1\%, respectively (Table~\ref{tab:testing_comparison_segmentation}). However, the optimal finetuning configuration varies across evaluation metrics. On AutoPET-III, Full FT achieves the lowest FPVol (15.594 cc), while Dec. FT achieves the lowest FNVol (7.405 cc). On Deep-PSMA, Dec. FT achieves the lowest FPVol (48.432 cc) and FNVol (6.588 cc). Despite these improvements, nnU-Net trained from scratch achieves the highest lesion-level PPV on both AutoPET-III (0.733) and Deep-PSMA (0.768), indicating a trade-off between lesion detection sensitivity and precision. We further present tracer-stratified Dice, FPVol, and FNVol in Figures~\ref{fig:autopet_tracer} and~\ref{fig:deep_psma_tracer}. For PSMA scans in AutoPET-III, Enc.~+~Dec. FT achieves the lowest FPVol (17.68 cc), whereas Dec. FT achieves the lowest FNVol (7.15 cc). In contrast, for PSMA scans in Deep-PSMA, Dec. FT achieves the lowest FPVol (57.73 cc), while Full FT achieves the lowest FNVol (4.90 cc). Overall, these results demonstrate that \textbf{LoRe-MIM} pretraining improves segmentation overlap across both datasets, while different finetuning configurations exhibit distinct trade-offs between false-positive and false-negative predictions.

\begin{table}[!htbp]
\centering
\caption{Lesion segmentation performance on the
AutoPET-III ($n=321$) and Deep-PSMA ($n=200$) test sets using different
baseline architectures and training strategies.
Three finetuning variants of LoRe-MIM (Ours) are evaluated:
Full FT, Dec. FT, and Enc. + Dec. FT.
FPVol and FNVol are reported in cc.
$\uparrow$: higher is better;
$\downarrow$: lower is better.}
\label{tab:testing_comparison_segmentation}


\begin{tabular}{c c c c c c c c c}
\toprule

\textbf{Dataset}
& \textbf{Baseline}
& \makecell{\textbf{Training}\\\textbf{Strategy}}
& \textbf{Variant}
& \textbf{Dice}$\uparrow$
& \textbf{FPVol}$\downarrow$
& \textbf{FNVol}$\downarrow$
& \makecell{\textbf{Lesion}\\\textbf{Sens.}$\uparrow$}
& \textbf{PPV}$\uparrow$ \\

\midrule

\multirow{5}{*}{AutoPET-III}
& nnU-Net
& Scratch & --
& 0.566 & 17.283 & 9.472
& 0.773 & \textbf{0.733} \\

\cmidrule(lr){2-9}

& \multirow{4}{*}{SwinUNETR}
& Scratch & --
& 0.526 & 27.147 & 7.586
& 0.773 & 0.546 \\

& & \multirow{3}{*}{\textbf{LoRe-MIM}}
& Full FT
& 0.577 & \textbf{15.594} & 9.832
& 0.775 & 0.664 \\

& & & Dec. FT
& 0.569 & 19.207 & \textbf{7.405}
& 0.764 & 0.713 \\

& & & Enc. + Dec. FT
& \textbf{0.581} & 15.662 & 8.362
& \textbf{0.776} & 0.675 \\

\midrule

\multirow{5}{*}{Deep-PSMA}
& nnU-Net
& Scratch & --
& 0.561 & 73.030 & 7.933
& 0.757 & \textbf{0.768} \\

\cmidrule(lr){2-9}

& \multirow{4}{*}{SwinUNETR}
& Scratch & --
& 0.547 & 74.315 & 9.785
& 0.711 & 0.732 \\

& & \multirow{3}{*}{\textbf{LoRe-MIM}}
& Full FT
& 0.572 & 76.037 & 8.964
& 0.731 & 0.766 \\

& & & Dec. FT
& 0.610 & \textbf{48.432} & \textbf{6.588}
& 0.743 & 0.765 \\

& & & Enc. + Dec. FT
& \textbf{0.613} & 48.456 & 7.023
& \textbf{0.759} & 0.741 \\

\bottomrule
\end{tabular}%
\end{table}

\begin{figure}[htbp]

  \centering
  \includegraphics[width=\linewidth]{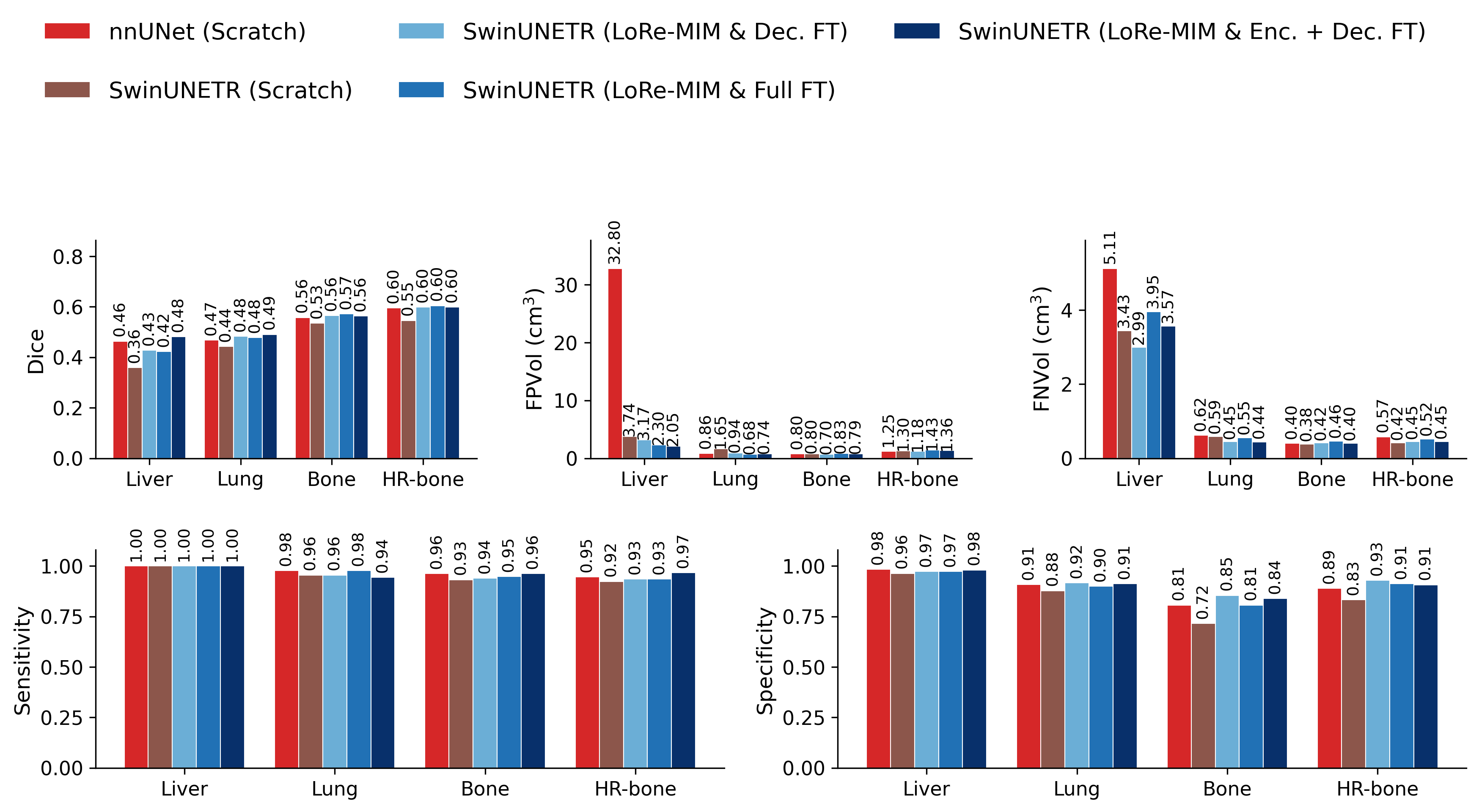}
  \caption{%
Anatomical region-level comparison of lesion segmentation performance on AutoPET-III, comparing nnU-Net and SwinUNETR trained from scratch with three finetuning variants of our \textbf{LoRe-MIM} pretrained SwinUNETR. 
}
  \label{fig:anatomy_autopet}
\end{figure}

\begin{figure}[htbp]

  \centering
  \includegraphics[width=\linewidth]{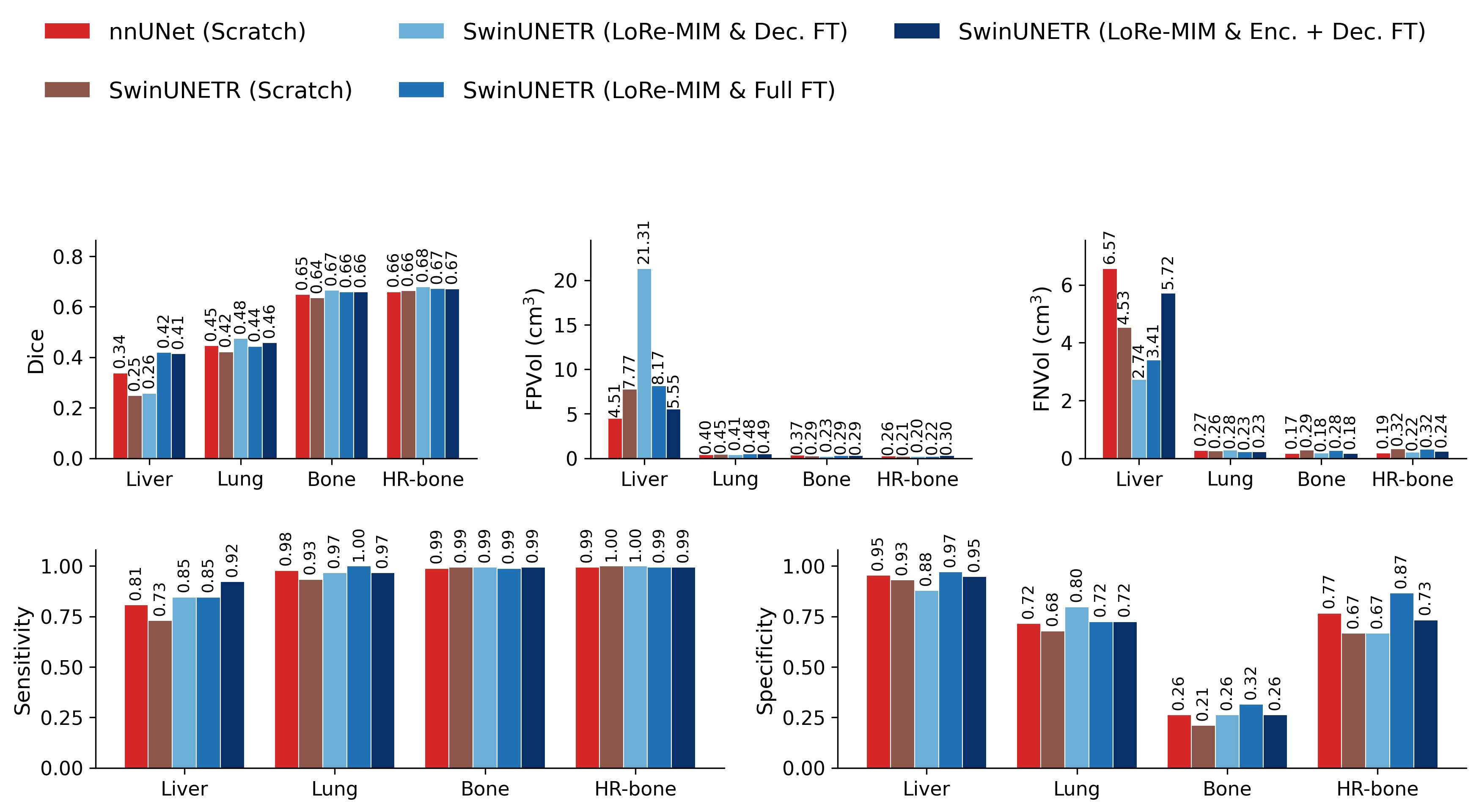}
  \caption{%
Anatomical region-level comparison of lesion segmentation performance on Deep PSMA, comparing nnU-Net and SwinUNETR trained from scratch with three finetuning variants of our \textbf{LoRe-MIM} pretrained SwinUNETR.
}
  \label{fig:anatomy_deep_psma}
\end{figure}

\paragraph{Anatomical region-level evaluation:} Figures~\ref{fig:anatomy_autopet} and~\ref{fig:anatomy_deep_psma} present anatomical-region-level segmentation performance across four regions (liver, lung, bone, and HR-bone) on AutoPET-III and Deep-PSMA, respectively, evaluated using Dice, FPVol, FNVol, lesion sensitivity, and specificity, computed for each region independently. Across both datasets, the best-performing \textbf{LoRe-MIM} finetuning variants consistently improve Dice over SwinUNETR trained from scratch in all four anatomical regions, with the largest improvements observed in the liver. Specifically, liver Dice increases from 0.36 to 0.48 on AutoPET-III and from 0.25 to 0.42 on Deep-PSMA, corresponding to relative improvements of 33.3\% and 68.0\%, respectively. Improvements are also observed in the lung (0.44 to 0.49 and 0.42 to 0.48), bone (0.53 to 0.57 and 0.64 to 0.67), and HR-bone (0.55 to 0.60 and 0.66 to 0.68) on AutoPET-III and Deep-PSMA, respectively. In addition to improved segmentation overlap, LoRe-MIM enhances lesion-level specificity in bone and HR-bone regions. On AutoPET-III, bone specificity increases from 0.72 to 0.85, while on Deep-PSMA, HR-bone specificity increases from 0.67 to 0.87 compared with SwinUNETR trained from scratch. For liver lesions, all three LoRe-MIM finetuning variants achieve higher sensitivity (0.85--0.92) than SwinUNETR trained from scratch (0.73) on Deep-PSMA, whereas liver sensitivity remains consistently high (1.00) across all configurations on AutoPET-III. Interestingly, improvements in liver Dice exhibit contrasting FPVol--FNVol trade-offs across the two datasets. On AutoPET-III, Enc.~+~Dec. FT reduces liver FPVol from 3.74 to 2.03 cc compared with SwinUNETR Scratch, accompanied by a slight increase in FNVol from 3.43 to 3.57 cc. Conversely, on Deep-PSMA, Dec. FT reduces liver FNVol from 4.53 to 2.74 cc while increasing FPVol from 7.77 to 21.31 cc. These findings suggest that \textbf{LoRe-MIM} improves anatomical-region-level segmentation across both datasets, although the relative contributions of reduced false-positive and false-negative volumes vary with the dataset and finetuning configuration.

\subsubsection{Label-efficiency curve}

\begin{figure}[htbp]

  \centering
  \includegraphics[width=\linewidth]{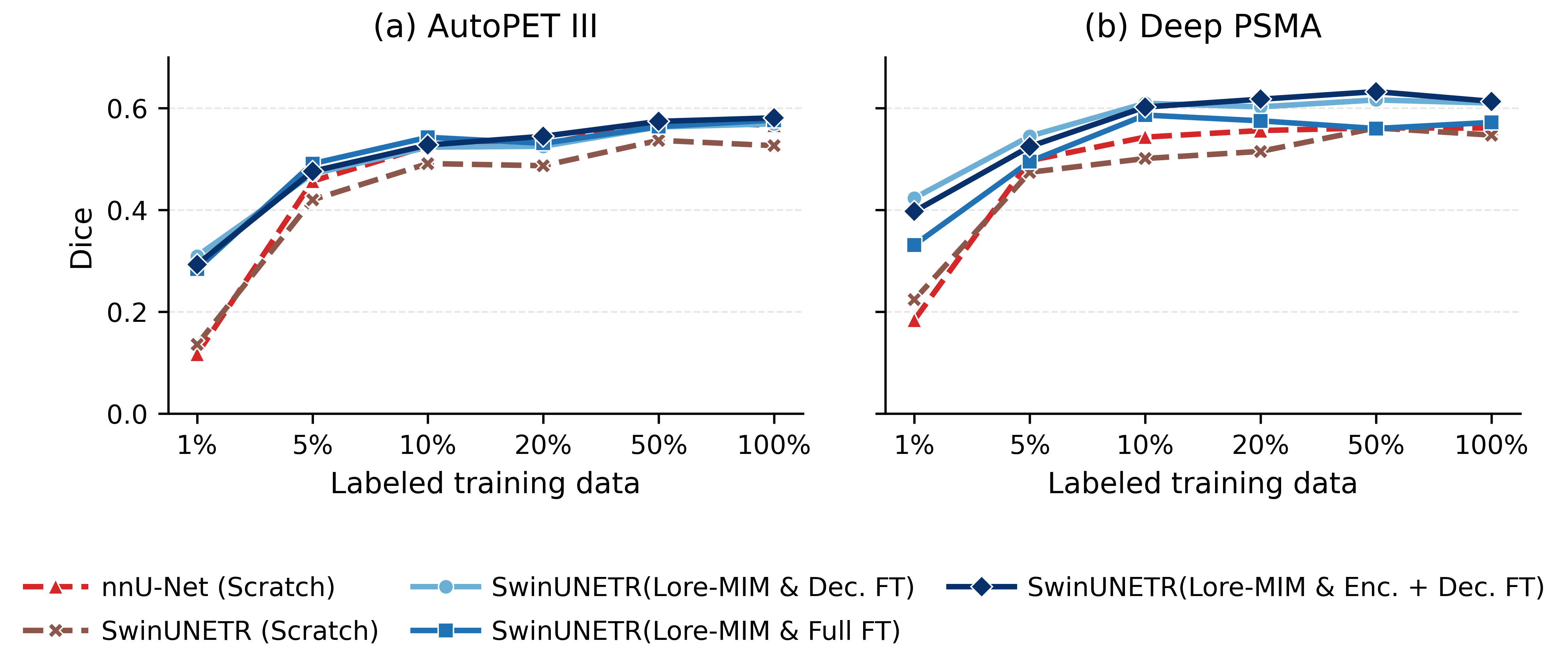}
\caption{Label-efficiency curves for whole-body lesion segmentation on the AutoPET-III and Deep-PSMA datasets. Segmentation performance is evaluated using Dice scores across varying proportions of labeled training data (1\%--100\%). All three finetuning strategies outperform SwinUNETR and nnU-Net trained from scratch, with more pronounced improvements under limited labeled data (1\%--10\%) and diminishing performance gains as the proportion of labeled data increases.}
  \label{fig:label_efficiency}
\end{figure}

In Figure~\ref{fig:label_efficiency}, we observe that under extremely limited labeled data (1\% and 5\%), our finetuning approaches achieve substantially higher Dice scores compared to their counterparts trained from scratch, with diminishing gains as the proportion of labeled data increases. These findings further support our motivation for developing a self-supervised pretraining approach that learns transferable representations for downstream tasks, especially under limited labeled data availability.

\subsubsection{Qualitative analysis}

\begin{figure}[htbp]

  \centering
  \includegraphics[width=\linewidth]{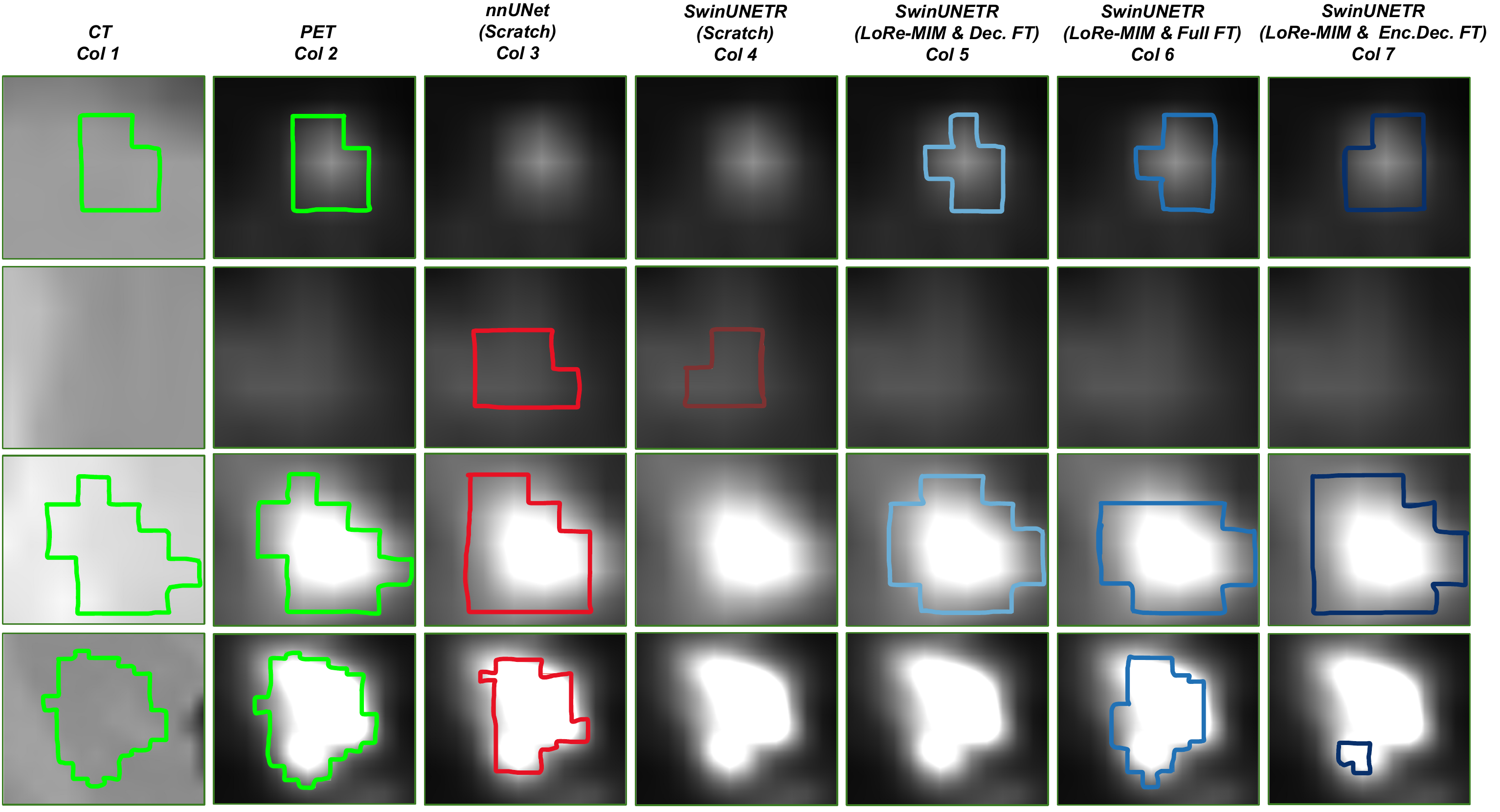}
\caption{
Qualitative comparison of lesion segmentation performance across four representative cases (Rows 1--4). Columns 1 and 2 show the CT and PET images with ground-truth lesion contours (green), respectively. Columns 3 and 4 present predictions from nnU-Net and SwinUNETR trained from scratch, respectively, while Columns 5--7 show predictions from three SwinUNETR finetuning strategies initialized with our pretrained \textbf{LoRe-MIM} model. 
Row 1 illustrates successful liver lesion recovery in an AutoPET-III case, where both scratch-trained models fail to detect the lesion, whereas all three LoRe-MIM variants successfully identify it.
Row 2 demonstrates false-positive elimination in an AutoPET-III case, where both scratch-trained models incorrectly predict a lesion, while all three LoRe-MIM variants correctly produce no lesion segmentation.
Row 3 presents a high-risk bone lesion from Deep-PSMA, where scratch-trained SwinUNETR misses the lesion, while all three LoRe-MIM variants successfully recover it.
Finally, Row 4 illustrates a failure case involving a liver lesion from Deep-PSMA, where decoder-only finetuning fails to detect the lesion and encoder–decoder finetuning yields an incomplete segmentation, whereas full finetuning and scratch-trained nnU-Net successfully recover the lesion.
}
  \label{fig:qual_lore_mim}
\end{figure}

\begin{figure}[htbp]

  \centering
  \includegraphics[width=\linewidth]{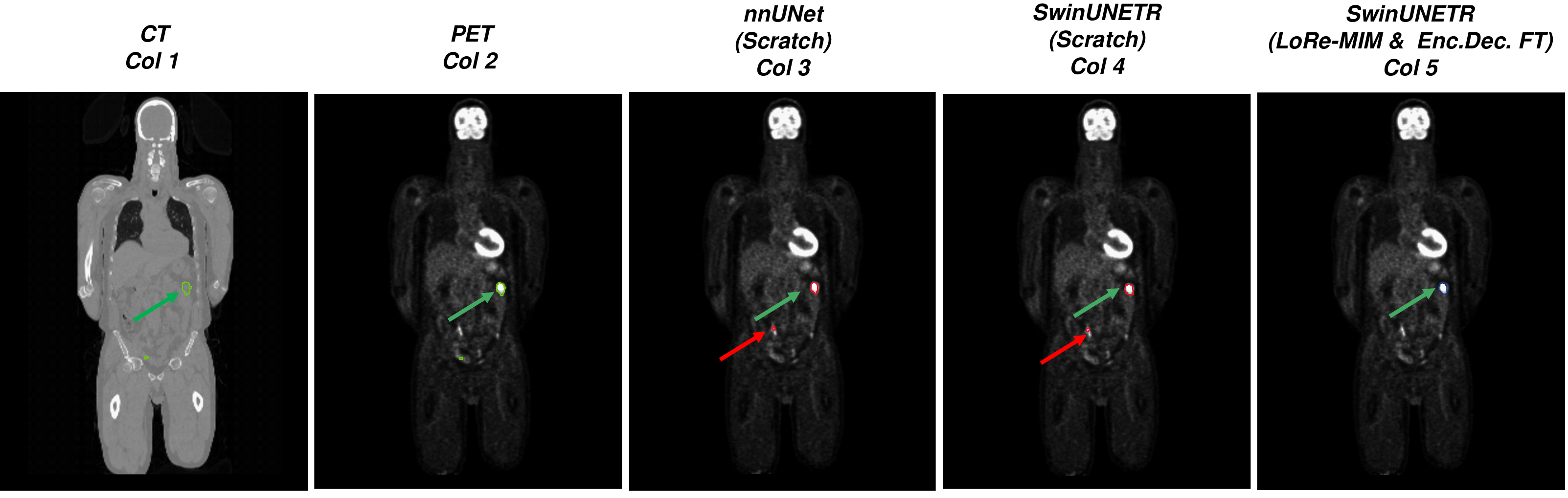}
\caption{
Qualitative comparison of whole-body PET/CT lesion segmentation between models trained from scratch (nnUNet and SwinUNETR) and our proposed LoRe-MIM pretrained SwinUNETR. Columns 1 and 2 show the coronal CT and PET images, respectively, with ground-truth lesion contours highlighted in green. Columns 3 and 4 present the segmentation predictions from nnUNet and SwinUNETR trained from scratch, respectively. Both models correctly identify the true-positive lesion (green arrow) but introduce an additional false-positive prediction (red arrow). In contrast, our proposed LoRe-MIM pretrained SwinUNETR (Column 5) successfully eliminates the false-positive prediction while preserving the true-positive lesion, demonstrating improved lesion discrimination. 
}
  \label{fig:qual_whole_body}
\end{figure}

We present two qualitative comparisons in Figures~\ref{fig:qual_lore_mim} and~\ref{fig:qual_whole_body}. Figure~\ref{fig:qual_lore_mim} illustrates representative examples of improved lesion recovery, false-positive elimination, and a failure case. Specifically, we select two cases from AutoPET-III (one liver lesion and one false-positive prediction) and two cases from Deep-PSMA (one liver lesion and one high-risk bone lesion). These examples demonstrate the effectiveness of our proposed LoRe-MIM finetuning variants in recovering lesions missed by models trained from scratch and suppressing false-positive predictions, while also highlighting their limitations in challenging cases. Furthermore, Figure~\ref{fig:qual_whole_body} presents a whole-body coronal visualization demonstrating that our proposed \textbf{LoRe-MIM} eliminates a false-positive prediction observed in both scratch-trained baselines (nnUNet and SwinUNETR) while preserving the true-positive lesion.

\begin{figure*}[t]
    \centering
    \includegraphics[width=\textwidth]
    {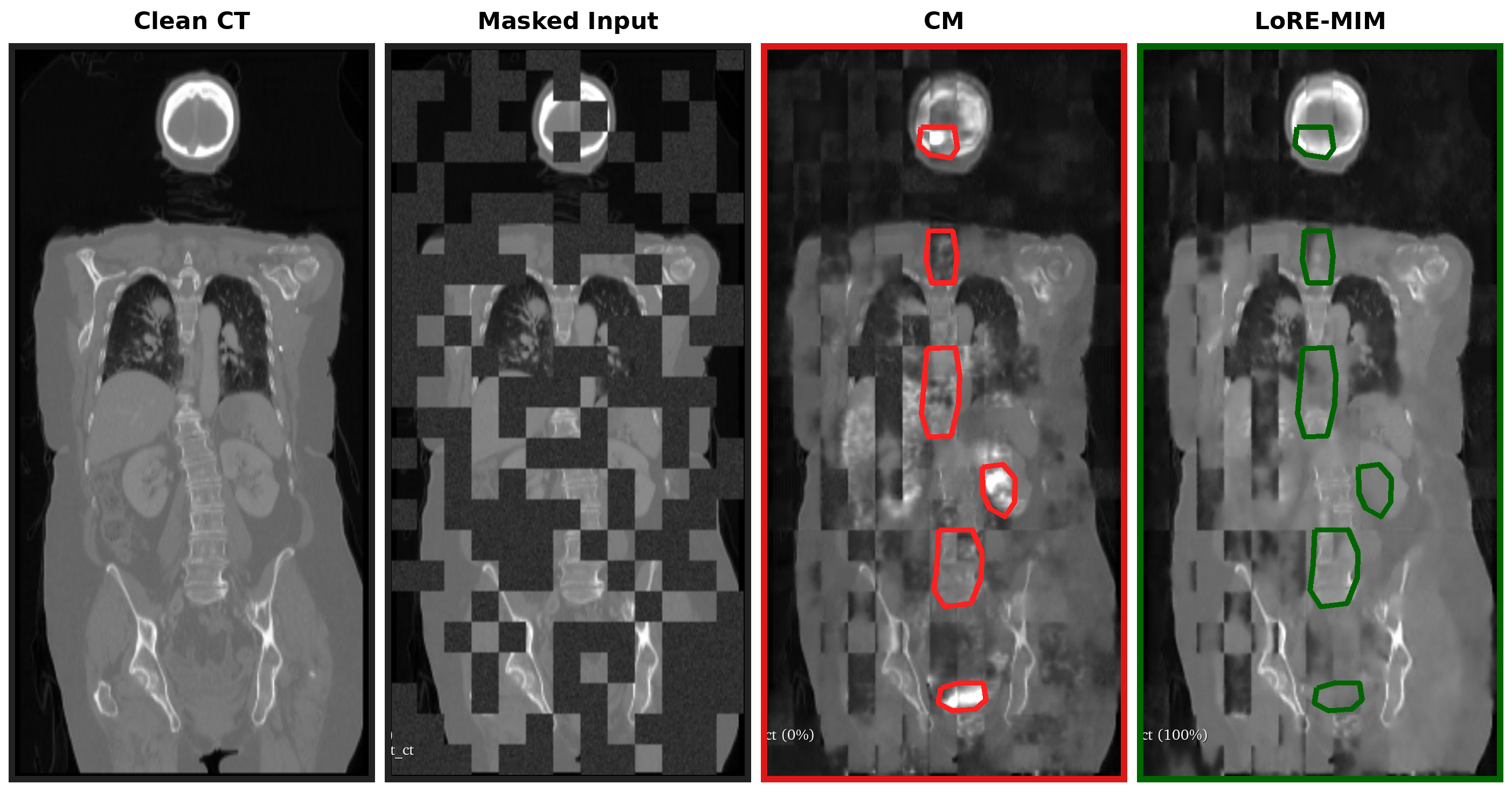}

    \vspace{3mm}

    \includegraphics[width=\textwidth]
    {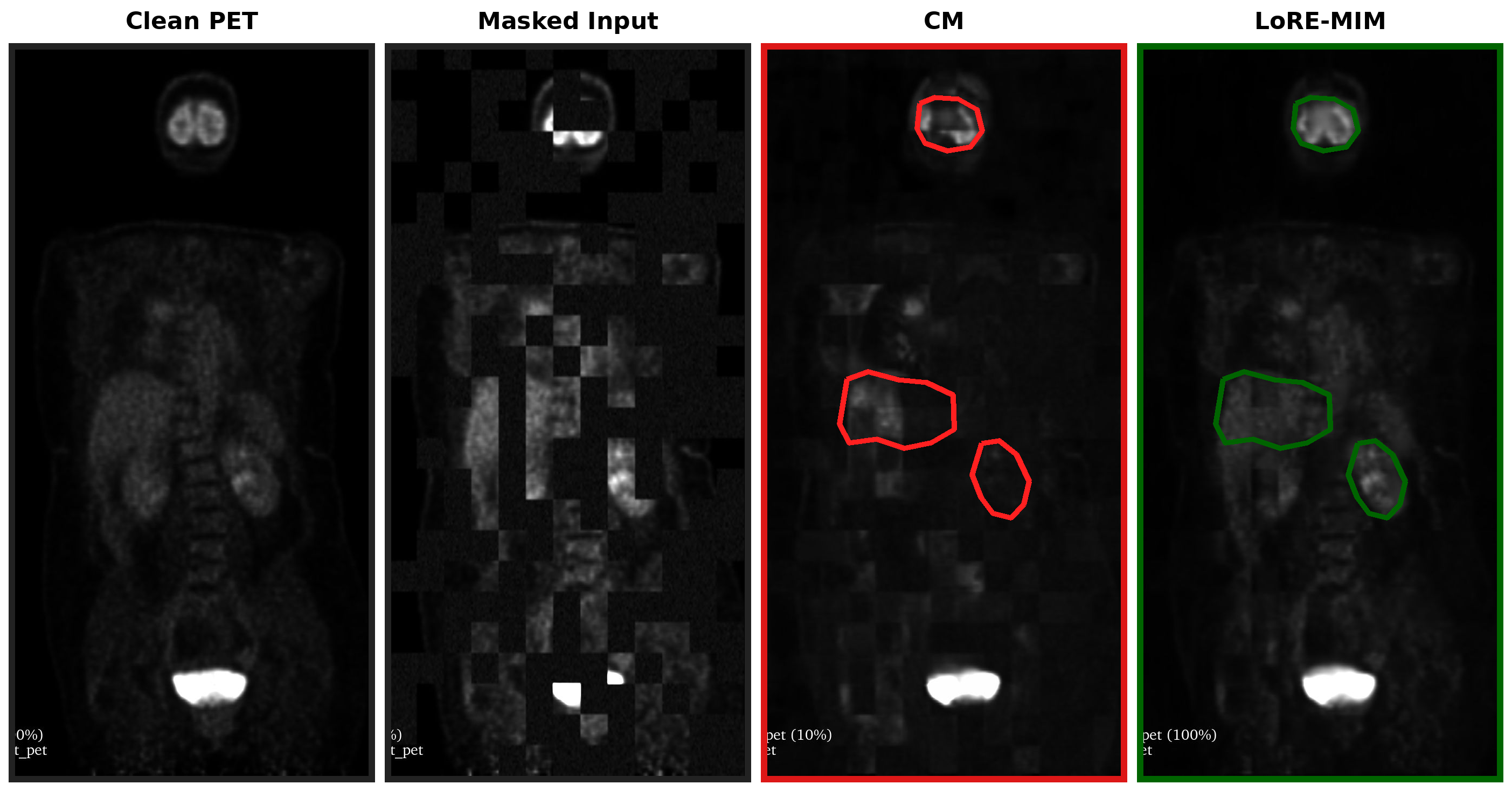}

    \caption{Qualitative comparison of masked-image reconstruction
    for CT (top) and PET (bottom). From left to right:
    clean reference, masked input, CM reconstruction, and
    LoRE-MIM reconstruction. Red region outlines highlight
    corresponding regions where LoRE-MIM provides visually
    closer agreement with the clean reference. CM and LoRE-MIM
    reconstructions are outlined in red and dark green,
    respectively.}
    \label{fig:qualitative_reconstruction}
\end{figure*}

\section{Discussion}

In this work, we aim to bridge the research gap arising from label scarcity in developing deep learning models for whole-body lesion segmentation and patch-based classification using PET/CT scans. To address this challenge, we seek to leverage well-established self-supervised pretraining schemes to learn rich and transferable representations of whole-body PET/CT scans. Existing pretraining approaches in the whole-body PET/CT domain often rely on relatively small cohorts, incorporate limited cross-modal interactions, and primarily evaluate the effectiveness of pretraining through downstream segmentation tasks only. In this work, we aim to address these limitations through a large-scale, heterogeneous pretraining dataset, a novel cross-modal masked image modeling strategy, and an extensive evaluation across multiple downstream tasks.

First, we introduce one of the largest cohorts for PET/CT pretraining, comprising $5910$ unlabeled scans collected from both publicly available and institutional datasets (Sect.~\ref{sec:datasets}). In contrast to most prior pretraining cohorts that are either only FDG- or PSMA-specific, our dataset incorporates pan-cancer, multi-tracer PET/CT scans spanning diverse patient populations and scanner variations. Such heterogeneity enables the pretraining framework to learn more generalizable and robust representations across different imaging conditions and disease characteristics. We investigate the generalizability of the learned representations through whole-body lesion segmentation across independent testing cohorts, while evaluating their robustness through patch-based classification under clinically relevant perturbations.

Second, we adopt a masked image modeling (MIM) based reconstruction objective, inspired by its success in natural images and medical imaging domains. However, the application of MIM in multimodal medical imaging remains relatively underexplored, and the few existing approaches often mimic unimodal masking strategies that capture relationships between spatially aligned imaging modalities and enable accurate prediction even when one modality is missing. This leaves considerable room for exploring cross-modal representation learning. To address this limitation, we propose \textbf{LoRe-MIM}, a novel masking strategy that emulates solving two puzzles simulataneously, in which some puzzle pieces from each puzzle are either lost or replaces with the alternate puzzle  (Sect.~\ref{sec:method}). Instead of adopting a fixed strategy for either removing or replacing all the masked blocks, we randomize the perturbation process using a Bernoulli random variable with a probability of 0.5. Through this stochastic masking mechanism, we encourage the pretrained model to learn anatomy and within-modal structural consistency as well as cross-modal relationships, correlations and complementarity, thereby capturing richer representations of multimodal PET/CT scans. Through ablation studies, we demonstrate that our proposed \textbf{LoRe-MIM} achieves the best reconstruction performance across all three metrics (MAE, SSIM, and PSNR), along with the highest classification AUC, compared to alternative masking strategies.

Third, we provide an extensive evaluation of our proposed dataset and pretraining strategy across multiple downstream tasks. While existing self-supervised pretraining approaches primarily evaluate their effectiveness through lesion segmentation, we investigate two complementary tasks: (1) patch-based classification and (2) whole-body lesion segmentation. We evaluate classification performance under both clean and perturbed imaging conditions, whereas segmentation performance is assessed at the voxel, lesion, and anatomical region levels. 

In patch-based classification, we systematically evaluate different feature extraction stages, feature aggregation strategies, and dimensionality reduction techniques. We observe that, among the evaluated configurations using encoder features, global average pooling (GAP) combined with logistic regression (LR) achieves the best classification performance. Interestingly, when adopting GAP and LR, features extracted from the decoder stages outperform those extracted from the encoder. Finally, using features extracted from the final decoder stage, we demonstrate that our pretrained representations achieve superior classification performance compared to widely adopted pretrained architectures, including DINOv3 and ResNet50-3D, under comparable feature dimensionality reduction settings.

We also benchmark the robustness of our pretrained representations under clinically relevant perturbations. Specifically, we introduce Gaussian noise into the 3D PET/CT patches, emulate spatial misalignment between PET and CT, and simulate missing modalities by independently removing either CT or PET. Across these perturbation settings, we observe that features extracted from the final decoder stage consistently outperform those obtained from DINOv3 and ResNet50-3D. These findings further support our hypothesis that the heterogeneous pretraining dataset and proposed LoRe-MIM strategy facilitate the learning of robust representations that maintain discriminative performance even under degraded or incomplete imaging conditions. Notably, the superior performance compared to DINOv3 highlights the potential of domain-specific multimodal pretraining over widely adopted general-purpose foundation models for PET/CT-based analysis.

Finally, we demonstrate the effectiveness of our pretraining approach for whole-body lesion segmentation. We investigate three different finetuning strategies by selectively finetuning different components of the pretrained model. Across all three strategies, for both of our testing datasets AutoPET-III and Deep-PSMA, we observe consistent improvements in Dice scores compared to the corresponding model trained from scratch, along with reductions in FPVol and FNVol in several configurations. Specifically, on AutoPET-III, finetuning both the encoder and decoder improves Dice from 0.526 to 0.581 and reduces FPVol from 27.147 cc to 15.662 cc, whereas on Deep-PSMA, decoder finetuning improves Dice from 0.547 to 0.610 while reducing FPVol from 74.315 cc to 48.432 cc and FNVol from 9.785 cc to 6.588 cc (Table~\ref{tab:testing_comparison_segmentation}). We further compare our approach with an independent nnU-Net architecture optimized for AutoPET-III, trained under comparable preprocessing, augmentation, and training conditions. Although our chosen architecture, SwinUNETR, trained from scratch performs inferiorly to nnU-Net, our pretraining strategy enables SwinUNETR to achieve improved segmentation performance and outperform the nnU-Net baseline in terms of Dice, achieving 0.581 compared to 0.566 on AutoPET-III and 0.613 compared to 0.561 on Deep-PSMA. Beyond voxel-level evaluation, we conduct comprehensive lesion-level and anatomical region-level analyses and observe that our pretraining approach improves lesion detection sensitivity and segmentation accuracy in clinically critical regions, particularly the liver and high-risk bone regions. These improvements highlight the potential clinical relevance of pretraining, especially for detecting and segmenting metastases in anatomically challenging and high-risk regions. Our label-efficiency curves in Figure~\ref{fig:label_efficiency} further demonstrate that the benefits of pretraining are most pronounced when only a limited number of labeled training samples are available, with diminishing gains as the number of labeled samples increases. The improvements in voxel- and lesion-level performance on the unseen Deep-PSMA dataset, which was not introduced during either pretraining or finetuning (Table~\ref{tab:testing_comparison_segmentation}), together with the anatomical region-level improvements (Figure~\ref{fig:anatomy_deep_psma}) and label-efficiency gains (Figure~\ref{fig:label_efficiency}), demonstrate the generalizability of the learned representations and highlight the effectiveness of our heterogeneous pretraining dataset and proposed perturbation strategy. 

Overall, our findings demonstrate that the combination of a large-scale, heterogeneous PET/CT dataset and the proposed LoRe-MIM masking strategy enables effective learning of generalizable and robust multimodal representations. Through extensive evaluation of reconstruction, classification, robustness, and whole-body lesion segmentation, we show that the benefits of self-supervised pretraining extend beyond improvements in individual downstream tasks. In particular, our results highlight the importance of incorporating cross-modal interactions during pretraining and leveraging heterogeneous imaging cohorts to develop more reliable and transferable deep learning models for whole-body PET/CT analysis.

\section*{Conclusion}

In this work, we introduced a novel self-supervised masked image modeling framework, $\textbf{LoRe-MIM}$, to learn generalizable and robust representations from whole-body PET/CT scans. We curated and subsequently leveraged a large-scale, heterogeneous pretraining cohort of 5,910 unlabeled PET/CT scans collected from both publicly available and institutional datasets. Our proposed block-based perturbation strategy encourages the learning of metabolic and complementary anatomical representations through a stochastic masking mechanism, in which image blocks are either removed (Lost) by replacing them with zero values or replaced (Replaced) with corresponding blocks from the alternate modality. We extensively evaluated our proposed framework across whole-body image reconstruction, patch-based classification of cancerous and non-cancerous regions, and whole-body lesion segmentation. We demonstrated that our method achieves superior reconstruction performance compared to an FDG-only pretrained model, improves classification performance under both clean and perturbed imaging conditions compared to established pretrained architectures such as DINOv3 and ResNet50-3D, and enhances lesion segmentation performance at the voxel, lesion, and anatomical region levels. Overall, these findings establish the utility of large-scale, heterogeneous pretraining datasets and the effectiveness of our proposed cross-modal self-supervised learning strategy in learning rich, generalizable, and robust representations for whole-body PET/CT analysis.

\section*{Acknowledgements}
Research reported in this publication was supported by an Institutional Development Award (IDeA) from the National Institute of General Medical Sciences of the National Institutes of Health under grant number 1P30GM149408. We also gratefully acknowledge support from the Munck-Pfefferkorn Fund, the American Cancer Society Institutional Research Grant, the Department of Biomedical Data Science at Dartmouth College and the Department of Radiology at Dartmouth Hitchcock Medical Center. All of this support was instrumental in making this work possible.
Help from Claude Opus 4.8 was taken for correcting grammatical errors.

\bibliography{example} 

@article{trotter2023positron,
  author  = {Trotter, Jacob and Pantel, Austin R. and Teo, Boon-Keng Kevin and Escorcia, Freddy E. and Li, Taoran and Pryma, Daniel A. and Taunk, Neil K.},
  title   = {Positron Emission Tomography ({PET})/Computed Tomography ({CT}) Imaging in Radiation Therapy Treatment Planning: A Review of {PET} Imaging Tracers and Methods to Incorporate {PET/CT}},
  journal = {Advances in Radiation Oncology},
  volume  = {8},
  number  = {5},
  pages   = {101212},
  year    = {2023},
  doi     = {10.1016/j.adro.2023.101212},
  url     = {https://doi.org/10.1016/j.adro.2023.101212}
}

@article{hofman2016we,
  author  = {Hofman, Michael S. and Hicks, Rodney J.},
  title   = {How We Read Oncologic {FDG PET/CT}},
  journal = {Cancer Imaging},
  volume  = {16},
  number  = {1},
  pages   = {35},
  year    = {2016},
  doi     = {10.1186/s40644-016-0091-3},
  url     = {https://doi.org/10.1186/s40644-016-0091-3}
}

@article{mingels2025total,
  author  = {Mingels, Clemens and Chung, Kevin J. and Pantel, Austin R. and Rominger, Axel and Alberts, Ian and Spencer, Benjamin A. and Nardo, Lorenzo and Pyka, Thomas},
  title   = {Total-Body {PET/CT}: Challenges and Opportunities},
  journal = {Seminars in Nuclear Medicine},
  volume  = {55},
  number  = {1},
  pages   = {21--30},
  year    = {2025},
  doi     = {10.1053/j.semnuclmed.2024.08.003},
  url     = {https://doi.org/10.1053/j.semnuclmed.2024.08.003}
}

@article{belal2024applications,
  author  = {Lindgren Belal, Sarah and Frantz, Sophia and Minarik, David and Enqvist, Olof and Wikstr{\"o}m, Erik and Edenbrandt, Lars and Tr{\"a}g{\aa}rdh, Elin},
  title   = {Applications of Artificial Intelligence in {PSMA PET/CT} for Prostate Cancer Imaging},
  journal = {Seminars in Nuclear Medicine},
  volume  = {54},
  number  = {1},
  pages   = {141--149},
  year    = {2024},
  doi     = {10.1053/j.semnuclmed.2023.06.001},
  url     = {https://doi.org/10.1053/j.semnuclmed.2023.06.001}
}

@article{gatidis2022fdg,
  author  = {Gatidis, Sergios and Hepp, Tobias and Fr{\"u}h, Marcel and La Foug{\`e}re, Christian and Nikolaou, Konstantin and Pfannenberg, Christina and Sch{\"o}lkopf, Bernhard and K{\"u}stner, Thomas and Cyran, Clemens and Rubin, Daniel},
  title   = {A Whole-Body {FDG-PET/CT} Dataset with Manually Annotated Tumor Lesions},
  journal = {Scientific Data},
  volume  = {9},
  number  = {1},
  pages   = {601},
  year    = {2022},
  doi     = {10.1038/s41597-022-01718-3},
  url     = {https://doi.org/10.1038/s41597-022-01718-3}
}

@misc{ingrisch2024autopet,
  author       = {Ingrisch, Michael and Dexl, Jakob and Jeblick, Katharina and Cyran, Clemens and Gatidis, Sergios and K{\"u}stner, Thomas},
  title        = {Automated Lesion Segmentation in Whole-Body {PET/CT} -- Multitracer Multicenter Generalization},
  year         = {2024},
  howpublished = {Challenge proposal, 27th International Conference on Medical Image Computing and Computer Assisted Intervention (MICCAI 2024)},
  publisher    = {Zenodo},
  doi          = {10.5281/zenodo.10990932},
  url          = {https://doi.org/10.5281/zenodo.10990932}
}

@article{rokuss2024fdg,
  author  = {Rokuss, Maximilian and Kovacs, Balint and Kirchhoff, Yannick and Xiao, Shuhan and Ulrich, Constantin and Maier-Hein, Klaus H. and Isensee, Fabian},
  title   = {From {FDG} to {PSMA}: A Hitchhiker's Guide to Multitracer, Multicenter Lesion Segmentation in {PET/CT} Imaging},
  journal = {arXiv preprint arXiv:2409.09478},
  year    = {2024},
  doi     = {10.48550/arXiv.2409.09478},
  url     = {https://arxiv.org/abs/2409.09478}
}

@article{yazdani2024automated,
  author  = {Yazdani, Elmira and Karamzadeh-Ziarati, Najme and Cheshmi, Seyyed Saeid and Sadeghi, Mahdi and Geramifar, Parham and Vosoughi, Habibeh and Kazemi Jahromi, Mahmood and Kheradpisheh, Saeed Reza},
  title   = {Automated Segmentation of Lesions and Organs at Risk on {[${}^{68}$Ga]Ga-PSMA-11} {PET/CT} Images Using Self-Supervised Learning with {Swin UNETR}},
  journal = {Cancer Imaging},
  volume  = {24},
  number  = {1},
  pages   = {30},
  year    = {2024},
  doi     = {10.1186/s40644-024-00675-x},
  url     = {https://doi.org/10.1186/s40644-024-00675-x}
}

@inproceedings{huang2025c2maot,
  author    = {Huang, Jiaju and Chen, Shaobin and Liang, Xinglong and Yang, Xiao and Zhang, Zhuoneng and Sun, Yue and Wang, Ying and Tan, Tao},
  title     = {{$\mathrm{C}^{2}$MAOT}: Cross-modal Complementary Masked Autoencoder with Optimal Transport for Cancer Segmentation in {PET-CT} Images},
  booktitle = {Medical Image Computing and Computer Assisted Intervention -- MICCAI 2025},
  series    = {Lecture Notes in Computer Science},
  volume    = {15960},
  pages     = {87--97},
  publisher = {Springer Nature Switzerland},
  year      = {2025},
  doi       = {10.1007/978-3-032-04927-8_9},
  url       = {https://doi.org/10.1007/978-3-032-04927-8_9}
}

@article{oh2025fratmae,
  author  = {Oh, Yujin and Seifert, Robert and Cao, Yihan and Clement, Christoph and Ferdinandus, Justin and Lapa, Constantin and Liebich, Alessandro and Amon, Michelle and Enke, Johanna and Song, Sifan and Meng, Runqi and Zeng, Fang and Guo, Ning and Li, Xiang and Heidari, Pedram and Rominger, Axel and Shi, Kuangyu and Li, Quanzheng},
  title   = {Developing a {PET/CT} Foundation Model for Cross-Modal Anatomical and Functional Imaging},
  journal = {arXiv preprint arXiv:2503.02824},
  year    = {2025},
  doi     = {10.48550/arXiv.2503.02824},
  url     = {https://arxiv.org/abs/2503.02824}
}

@article{liu2026open,
  author  = {Liu, Xiaofeng and Zhang, Qianru and Marin, Thibault and Xia, Menghua and Liu, Chi and El Fakhri, Georges and Ouyang, Jinsong},
  title   = {An Open Multi-Center Whole-Body {FDG PET/CT} Foundation Model for Tumor Segmentation},
  journal = {arXiv preprint arXiv:2605.21835},
  year    = {2026},
  doi     = {10.48550/arXiv.2605.21835},
  url     = {https://arxiv.org/abs/2605.21835}
}

@inproceedings{swinunetr,
  author    = {Hatamizadeh, Ali and Nath, Vishwesh and Tang, Yucheng and Yang, Dong and Roth, Holger R. and Xu, Daguang},
  title     = {Swin {UNETR}: Swin Transformers for Semantic Segmentation of Brain Tumors in {MRI} Images},
  booktitle = {Brainlesion: Glioma, Multiple Sclerosis, Stroke and Traumatic Brain Injuries},
  series    = {Lecture Notes in Computer Science},
  volume    = {12962},
  pages     = {272--284},
  publisher = {Springer International Publishing},
  address   = {Cham},
  year      = {2022},
  doi       = {10.1007/978-3-031-08999-2_22},
  url       = {https://doi.org/10.1007/978-3-031-08999-2_22}
}

@dataset{jackson2025deeppsma,
  author    = {Jackson, Price and Hofman, Michael and Buteau, James Patrick
               and McIntosh, Lachlan and Sun, Yu},
  title     = {A Repository of Annotated {PSMA} and {FDG PET/CT} Images
               for Algorithm Development in Staging of {mCRPC} for
               Treatment with {$^{177}$Lu-PSMA} Therapy},
  year      = {2025},
  publisher = {Zenodo},
  doi       = {10.5281/zenodo.15281784},
  url       = {https://doi.org/10.5281/zenodo.15281784}
}

@article{eyuboglu2021weak,
  author  = {Eyuboglu, Sabri and Angus, Geoffrey and Patel, Bhavik N.
             and Pareek, Anuj and Davidzon, Guido and Long, Jin
             and Dunnmon, Jared and Lungren, Matthew P.},
  title   = {Multi-task weak supervision enables anatomically-resolved
             abnormality detection in whole-body {FDG-PET/CT}},
  journal = {Nature Communications},
  volume  = {12},
  number  = {1},
  pages   = {1880},
  year    = {2021},
  doi     = {10.1038/s41467-021-22018-1},
  url     = {https://doi.org/10.1038/s41467-021-22018-1}
}

@inproceedings{nguyen2025vietnamese,
  author    = {Nguyen, Huu Tien and Nguyen, Dac Thai
               and Nguyen, The Minh Duc and Nguyen, Trung Thanh
               and Truong, Thao Nguyen and Pham, Huy Hieu
               and Barthelemy, Johan and Tran, Minh Quan
               and Nguyen, Thanh Tam and Nguyen, Quoc Viet Hung
               and Chau, Quynh Anh and Mai, Hong Son
               and Nguyen, Thanh Trung and Nguyen, Phi Le},
  title     = {Toward a Vision-Language Foundation Model for Medical Data:
               Multimodal Dataset and Benchmarks for Vietnamese
               {PET/CT} Report Generation},
  booktitle = {Advances in Neural Information Processing Systems},
  volume    = {38},
  pages     = {129875--129906},
  year      = {2025},
  doi       = {10.52202/085713-3907},
  url       = {https://doi.org/10.52202/085713-3907}
}

@article{jemaa2020tumor,
  title={Tumor segmentation and feature extraction from whole-body FDG-PET/CT using cascaded 2D and 3D convolutional neural networks},
  author={Jemaa, Skander and Fredrickson, Jill and Carano, Richard AD and Nielsen, Tina and de Crespigny, Alex and Bengtsson, Thomas},
  journal={Journal of digital imaging},
  volume={33},
  number={4},
  pages={888--894},
  year={2020},
  publisher={Springer}
}

@inproceedings{milletari2016v,
  title={V-net: Fully convolutional neural networks for volumetric medical image segmentation},
  author={Milletari, Fausto and Navab, Nassir and Ahmadi, Seyed-Ahmad},
  booktitle={2016 fourth international conference on 3D vision (3DV)},
  pages={565--571},
  year={2016},
  organization={Ieee}
}

@article{xu2018automated,
  title={Automated whole-body bone lesion detection for multiple myeloma on 68Ga-pentixafor PET/CT imaging using deep learning methods},
  author={Xu, Lina and Tetteh, Giles and Lipkova, Jana and Zhao, Yu and Li, Hongwei and Christ, Patrick and Piraud, Marie and Buck, Andreas and Shi, Kuangyu and Menze, Bjoern H},
  journal={Contrast media \& molecular imaging},
  volume={2018},
  number={1},
  pages={2391925},
  year={2018},
  publisher={Wiley Online Library}
}

@article{blanc2021fully,
  title={Fully automatic segmentation of diffuse large B cell lymphoma lesions on 3D FDG-PET/CT for total metabolic tumour volume prediction using a convolutional neural network.},
  author={Blanc-Durand, Paul and J{\'e}gou, Simon and Kanoun, Salim and Berriolo-Riedinger, Alina and Bodet-Milin, Caroline and Kraeber-Bod{\'e}r{\'e}, Fran{\c{c}}oise and Carlier, Thomas and Le Gouill, Steven and Casasnovas, Ren{\'e}-Olivier and Meignan, Michel and others},
  journal={European Journal of Nuclear Medicine and Molecular Imaging},
  volume={48},
  number={5},
  pages={1362--1370},
  year={2021},
  publisher={Springer}
}

@inproceedings{liu2024lm,
  title={LM-UNet: Whole-Body PET-CT Lesion Segmentation with Dual-Modality-Based Annotations Driven by Latent Mamba U-Net},
  author={Liu, Anglin and Jia, Dengqiang and Sun, Kaicong and Meng, Runqi and Zhao, Meixin and Jiang, Yongluo and Dong, Zhijian and Gao, Yaozong and Shen, Dinggang},
  booktitle={International Conference on Medical Image Computing and Computer-Assisted Intervention},
  pages={405--414},
  year={2024},
  organization={Springer}
}

@article{gatidis2024results,
  title={Results from the autoPET challenge on fully automated lesion segmentation in oncologic PET/CT imaging},
  author={Gatidis, Sergios and Fr{\"u}h, Marcel and Fabritius, Matthias P and Gu, Sijing and Nikolaou, Konstantin and Foug{\`e}re, Christian La and Ye, Jin and He, Junjun and Peng, Yige and Bi, Lei and others},
  journal={Nature Machine Intelligence},
  volume={6},
  number={11},
  pages={1396--1405},
  year={2024},
  publisher={Nature Publishing Group UK London}
}

@article{he2023whole,
  title={Whole-body tumor segmentation from PET/CT images using a two-stage cascaded neural network with camouflaged object detection mechanisms},
  author={He, Jiangping and Zhang, Yangjie and Chung, Maggie and Wang, Michael and Wang, Kun and Ma, Yan and Ding, Xiaoyang and Li, Qiang and Pu, Yonglin},
  journal={Medical Physics},
  volume={50},
  number={10},
  pages={6151--6162},
  year={2023},
  publisher={Wiley Online Library}
}

@article{xu2023automatic,
  title={Automatic segmentation of prostate cancer metastases in PSMA PET/CT images using deep neural networks with weighted batch-wise dice loss},
  author={Xu, Yixi and Klyuzhin, Ivan and Harsini, Sara and Ortiz, Anthony and Zhang, Shun and B{\'e}nard, Fran{\c{c}}ois and Dodhia, Rahul and Uribe, Carlos F and Rahmim, Arman and Ferres, Juan Lavista},
  journal={Computers in Biology and Medicine},
  volume={158},
  pages={106882},
  year={2023},
  publisher={Elsevier}
}

@article{li2024automated,
  title={An automated deep learning-based framework for uptake segmentation and classification on PSMA PET/CT imaging of patients with prostate cancer},
  author={Li, Yang and Imami, Maliha R and Zhao, Linmei and Amindarolzarbi, Alireza and Mena, Esther and Leal, Jeffrey and Chen, Junyu and Gafita, Andrei and Voter, Andrew F and Li, Xin and others},
  journal={Journal of Imaging Informatics in Medicine},
  volume={37},
  number={5},
  pages={2206--2215},
  year={2024},
  publisher={Springer}
}

@article{keshavarz2026self,
author={Keshavarz, Sajad and Saeedzadeh, Elham and Arabi, Hossein and Jenabi-Haghparast, Elnaz and Sardari, Dariush},
  title={Self-supervised multimodal swin UNETR for PET/CT segmentation of diffuse large B-cell lymphoma},
  journal={Cancer Treatment and Research Communications},
  volume={49},
  pages={101369},
  year={2026},
  publisher={Elsevier}
}

@article{alves2024classification,
  title={Classification of pulmonary nodules in 2-[18F] FDG PET/CT images with a 3D convolutional neural network},
  author={Alves, Victor Manuel and dos Santos Cardoso, Jaime and Gama, Jo{\~a}o},
  journal={Nuclear Medicine and Molecular Imaging},
  volume={58},
  number={1},
  pages={9--24},
  year={2024},
  publisher={Springer}
}

@article{simonyan2014very,
  title={Very deep convolutional networks for large-scale image recognition},
  author={Simonyan, Karen and Zisserman, Andrew},
  journal={arXiv preprint arXiv:1409.1556},
  year={2014}
}

@article{callahan2011clinical,
  title={The clinical significance and management of lesion motion due to respiration during PET/CT scanning},
  author={Callahan, Jason and Kron, Tomas and Schneider-Kolsky, Michal and Hicks, Rodney J},
  journal={Cancer Imaging},
  volume={11},
  number={1},
  pages={224},
  year={2011}
}

@article{zaidi2007optimisation,
  title={Optimisation of whole-body PET/CT scanning protocols},
  author={Zaidi, H},
  journal={Biomedical imaging and intervention journal},
  volume={3},
  number={2},
  pages={e36},
  year={2007}
}

@article{osman2003respiratory,
  title={Respiratory motion artifacts on PET emission images obtained using CT attenuation correction on PET-CT},
  author={Osman, Medhat M and Cohade, Christian and Nakamoto, Yuji and Wahl, Richard L},
  journal={European journal of nuclear medicine and molecular imaging},
  volume={30},
  number={4},
  pages={603--606},
  year={2003},
  publisher={Springer}
}

@article{cardoso2022monai,
  title={MONAI: An open-source framework for deep learning in healthcare},
  author={Cardoso, M. Jorge and Li, Wenqi and Brown, Richard and Ma, Nhat and Kerfoot, Eric and Wang, Yiheng and Murrey, Benjamin and Myronenko, Andriy and Zhao, Can and Yang, Dong and others},
  journal={arXiv preprint arXiv:2211.02701},
  year={2022}
}

@inproceedings{hatamizadeh2021swin,
  title={Swin unetr: Swin transformers for semantic segmentation of brain tumors in mri images},
  author={Hatamizadeh, Ali and Nath, Vishwesh and Tang, Yucheng and Yang, Dong and Roth, Holger R and Xu, Daguang},
  booktitle={International MICCAI brainlesion workshop},
  pages={272--284},
  year={2021},
  organization={Springer}
}

@article{tao202118f,
  title={18F-FDG-PET/CT Whole-Body Imaging Lung Tumor Diagnostic Model: An Ensemble E-ResNet-NRC with Divided Sample Space},
  author={Tao, Zhou and Bing-Qiang, Huo and Huiling, Lu and Hongbin, Shi and Pengfei, Yang and Hongsheng, Ding},
  journal={BioMed Research International},
  volume={2021},
  number={1},
  pages={8865237},
  year={2021},
  publisher={Wiley Online Library}
}

@article{simeoni2025dinov3,
  title={Dinov3},
  author={Sim{\'e}oni, Oriane and Vo, Huy V and Seitzer, Maximilian and Baldassarre, Federico and Oquab, Maxime and Jose, Cijo and Khalidov, Vasil and Szafraniec, Marc and Yi, Seungeun and Ramamonjisoa, Micha{\"e}l and others},
  journal={arXiv preprint arXiv:2508.10104},
  year={2025}
}

@article{biswas2026must,
  title={MUST-PET: MUltimodal Self-supervised learning across Tracers for whole-body PET/CT-based lesion segmentation},
  author={Biswas, Bashirul Azam and Bhattacharya, Amartya and Wagle, Biratal Raj and Maeder, Matthew E and Yu, James B and Bhattacharya, Indrani},
  journal={arXiv preprint arXiv:2608.19666},
  year={2026}
}

@article{chen2019med3d,
  title={Med3d: Transfer learning for 3d medical image analysis},
  author={Chen, Sihong and Ma, Kai and Zheng, Yefeng},
  journal={arXiv preprint arXiv:1904.00625},
  year={2019}
}

@article{wasserthal2023totalsegmentator,
  title={TotalSegmentator: robust segmentation of 104 anatomic structures in CT images},
  author={Wasserthal, Jakob and Breit, Hanns-Christian and Meyer, Manfred T and Pradella, Maurice and Hinck, Daniel and Sauter, Alexander W and Heye, Tobias and Boll, Daniel T and Cyriac, Joshy and Yang, Shan and others},
  journal={Radiology: Artificial Intelligence},
  volume={5},
  number={5},
  pages={e230024},
  year={2023},
  publisher={Radiological Society of North America}
}

@article{sung2026global,
  title={Global cancer statistics 2024: GLOBOCAN estimates of incidence and mortality worldwide for 34 cancers in 186 countries},
  author={Sung, Hyuna and Filho, Adalberto M and Laversanne, Mathieu and Ferlay, Jacques and Siegel, Rebecca L and Soerjomataram, Isabelle and Jemal, Ahmedin and Bray, Freddie},
  journal={CA: A Cancer Journal for Clinicians},
  volume={76},
  number={4},
  pages={e70090},
  year={2026},
  publisher={Wiley Online Library}
}

@article{nishigaki2024vision,
  title={Vision transformer to differentiate between benign and malignant slices in 18F-FDG PET/CT},
  author={Nishigaki, Daiki and Suzuki, Yuki and Watabe, Tadashi and Katayama, Daisuke and Kato, Hiroki and Wataya, Tomohiro and Kita, Kosuke and Sato, Junya and Tomiyama, Noriyuki and Kido, Shoji},
  journal={Scientific Reports},
  volume={14},
  number={1},
  pages={8334},
  year={2024},
  publisher={Nature Publishing Group UK London}
}

@article{yang2023computer,
  title={Computer-aided diagnostic models to classify lymph node metastasis and lymphoma involvement in enlarged cervical lymph nodes using PET/CT},
  author={Yang, Yuhan and Zheng, Bo and Li, Yueyi and Li, Yuan and Ma, Xuelei},
  journal={Medical Physics},
  volume={50},
  number={1},
  pages={152--162},
  year={2023},
  publisher={Wiley Online Library}
}

@inproceedings{deng2009imagenet,
  title={Imagenet: A large-scale hierarchical image database},
  author={Deng, Jia and Dong, Wei and Socher, Richard and Li, Li-Jia and Li, Kai and Fei-Fei, Li},
  booktitle={2009 IEEE conference on computer vision and pattern recognition},
  pages={248--255},
  year={2009},
  organization={Ieee}
}

@inproceedings{he2016deep,
  title={Deep residual learning for image recognition},
  author={He, Kaiming and Zhang, Xiangyu and Ren, Shaoqing and Sun, Jian},
  booktitle={Proceedings of the IEEE conference on computer vision and pattern recognition},
  pages={770--778},
  year={2016}
}

@inproceedings{szegedy2016rethinking,
  title={Rethinking the inception architecture for computer vision},
  author={Szegedy, Christian and Vanhoucke, Vincent and Ioffe, Sergey and Shlens, Jon and Wojna, Zbigniew},
  booktitle={Proceedings of the IEEE conference on computer vision and pattern recognition},
  pages={2818--2826},
  year={2016}
}

@article{krizhevsky2012imagenet,
  title={Imagenet classification with deep convolutional neural networks},
  author={Krizhevsky, Alex and Sutskever, Ilya and Hinton, Geoffrey E},
  journal={Advances in neural information processing systems},
  volume={25},
  year={2012}
}

@article{sibille202018f,
  title={18F-FDG PET/CT uptake classification in lymphoma and lung cancer by using deep convolutional neural networks},
  author={Sibille, Ludovic and Seifert, Robert and Avramovic, Nemanja and Vehren, Thomas and Spottiswoode, Bruce and Zuehlsdorff, Sven and Sch{\"a}fers, Michael},
  journal={Radiology},
  volume={294},
  number={2},
  pages={445--452},
  year={2020},
  publisher={Radiological Society of North America}
}

@article{shao2024transfer,
  title={Transfer learning--based PET/CT three-dimensional convolutional neural network fusion of image and clinical information for prediction of EGFR mutation in lung adenocarcinoma},
  author={Shao, Xiaonan and Ge, Xinyu and Gao, Jianxiong and Niu, Rong and Shi, Yunmei and Shao, Xiaoliang and Jiang, Zhenxing and Li, Renyuan and Wang, Yuetao},
  journal={BMC Medical Imaging},
  volume={24},
  number={1},
  pages={54},
  year={2024},
  publisher={Springer}
}

@article{mu2020non,
  title={Non-invasive decision support for NSCLC treatment using PET/CT radiomics},
  author={Mu, Wei and Jiang, Lei and Zhang, JianYuan and Shi, Yu and Gray, Jhanelle E and Tunali, Ilke and Gao, Chao and Sun, Yingying and Tian, Jie and Zhao, Xinming and others},
  journal={Nature communications},
  volume={11},
  number={1},
  pages={5228},
  year={2020},
  publisher={Nature Publishing Group UK London}
}

@article{shao2021application,
  title={Application of dual-stream 3D convolutional neural network based on 18F-FDG PET/CT in distinguishing benign and invasive adenocarcinoma in ground-glass lung nodules},
  author={Shao, Xiaonan and Niu, Rong and Shao, Xiaoliang and Gao, Jianxiong and Shi, Yunmei and Jiang, Zhenxing and Wang, Yuetao},
  journal={EJNMMI physics},
  volume={8},
  number={1},
  pages={74},
  year={2021},
  publisher={Springer}
}
\bibliographystyle{spiebib} 

\end{document}